# Jointly RS Image Deblurring and Super-Resolution with Adjustable-Kernel and Multi-Domain Attention

Yan Zhang, Pengcheng Zheng†, Chengxiao Zeng, Bin Xiao, Zhenghao Li, and Xinbo Gao, *Fellow, IEEE*

*Abstract*—Remote Sensing (RS) image deblurring and Super-Resolution (SR) are common tasks in computer vision that aim at restoring RS image detail and spatial scale, respectively. However, real-world RS images often suffer from a complex combination of global low-resolution (LR) degeneration and local blurring degeneration. Although carefully designed deblurring and SR models perform well on these two tasks individually, a unified model that performs jointly RS image deblurring and super-resolution (JRSIDSR) task is still challenging due to the vital dilemma of reconstructing the global and local degeneration simultaneously. Additionally, existing methods struggle to capture the interrelationship between deblurring and SR processes, leading to suboptimal results. To tackle these issues, we give a unified theoretical analysis of RS images' spatial and blur degeneration processes and propose a dual-branch parallel network named AKMD-Net for the JRSIDSR task. AKMD-Net consists of two main branches: deblurring and super-resolution branches. In the deblurring branch, we design a pixel-adjustable kernel block (PAKB) to estimate the local and spatial-varying blur kernels. In the SR branch, a multi-domain attention block (MDAB) is proposed to capture the global contextual information enhanced with high-frequency details. Furthermore, we develop an adaptive feature fusion (AFF) module to model the contextual relationships between the deblurring and SR branches. Finally, we design an adaptive Wiener loss (AW Loss) to depress the prior noise in the reconstructed images. Extensive experiments demonstrate that the proposed AKMD-Net achieves state-of-the-art (SOTA) quantitative and qualitative performance on commonly used RS image datasets. The source code is publicly available at *https://github.com/zpc456/AKMD-Net*.

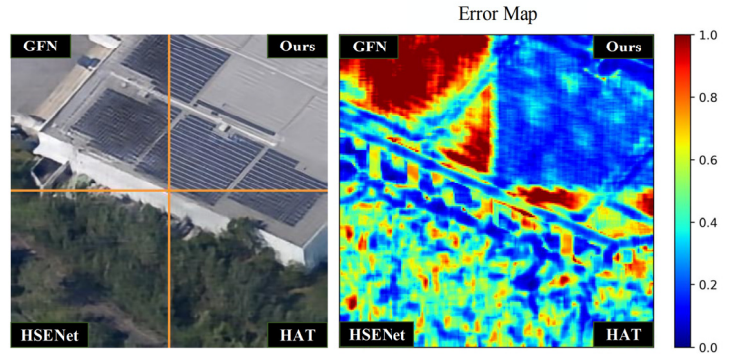

Fig. 1. Representative results of different models. Images reconstructed by our model obtain the best visual performance (as shown in the left part) and the least error map response (as shown in the right part).

## I. INTRODUCTION

This work was supported in part by the National Natural Science Foundation of China under Grant 62036007, Grant 62176195 and Grant 62206034; in part by the special support of Chongqing postdoctoral research project 2023CQB-SHTB3097; in part by the Special Project on Technological Innovation and Application Development under Grant cstc2020jscx-dxwtB0032; in part by Chongqing Excellent Scientist Project under Grant cstc2021ycjhbgzxm0339; in part by Natural Science Foundation of Chongqing under Grant cstc2021jcyj-msxmX0847; and in part by the Science and Technology Research Program of Chongqing Municipal Education Commission (Grant No. KJQN202200644 and Grant No. KJQN202200618). (†*Yan Zhang and Pengcheng Zheng are co-first authors.*) (*Corresponding author: Xinbo Gao.*)

Yan Zhang is with the Department of Computer Science and Technology, Chongqing University of Posts and Telecommunications, Chongqing 400065, China (e-mail: yanzhang1991@cqupt.edu.cn) and State Key Laboratory of Integrated Service Networks.

Pengcheng Zheng, Chengxiao Zeng, Bin Xiao and Xinbo Gao, are with the Department of Computer Science and Technology, Chongqing University of Posts and Telecommunications, Chongqing 400065, China (e-mail: 2020211905@stu.cqupt.edu.cn; 2021211912@stu.cqupt.edu.cn; xiaobin@cqupt.edu.cn; gaoxb@cqupt.edu.cn;).

Zhenghao Li is with the Chongqing Institute of Green and Intelligent Technology, Chinese Academy of Sciences, Chongqing 400713, China (email: lizh@cigit.ac.cn).

**B**ENEFITING from advancements in satellite imaging technology, the availability of remote-sensing (RS) images has significantly increased. RS images often exhibit lower noise levels compared to other types of images. Consequently, the jointly remote-sensing image deblurring and super-resolution (JRSIDSR) task aims to enhance the quality of a blurry and low-resolution image by removing blur and increasing its resolution. In contrast, the real-world image super-resolution (RWISR) task typically addresses more complex and realistic degradation models, which include noise, artifacts, and varying levels of blur. High-quality RS images are crucial for achieving superior results in various high-level vision tasks, such as RS semantic segmentation [1], change detection [2], and land use classification [3]. It is crucial for real-world demands and have widespread applications in practical settings, such as precision agriculture, disaster management, and urban planning [4, 5]. However, the complex imaging environments inherent to RS often result in degraded image quality, characterized by relative motion blur [6, 7], atmospheric turbulence [8], and low-resolution spatial degeneration [9]. Hence, how to handle jointly remote-sensing image deblurring and super-resolution (JRSIDSR) task has attracted significant attention from the RS community [10, 11].

In the era of deep learning, numerous methods [12, 13, 14] have been proposed to tackle jointly image deblurring and super-resolution (JIDSR) challenges, typically categorized into cascading and parallel paradigms. For the cascading paradigm, a deblurring branch is followed by a super-resolution branch, or vice versa. However, this approach is limited by its inability to fully leverage the contextual features of both deblurring and super-resolution tasks, often resulting in suboptimal results. Additionally, the cascading paradigm would accumulate the reconstructed errors [15], leading to suboptimal performance. Therefore, recent methods generally adopt the parallel paradigm for the JIDSR task. For example, the ED-DSRN [14] model applied a dual-branch encoder-decoder architecture [16] to exploit the multi-task features and obtain perceptually more convincing results. Nevertheless, this model primarily focuses on mitigating uniform Gaussian blur [17] and performs less well on other RS image degeneration. Thus, Zhang et al. [12] proposed a dual-branch convolutional neural network (GFN) to extract super-resolution and deblurring features and then merge them using the recursive gate module. However, these methods lack theoretical analysis of RS image degeneration process, which complicates the selection of an appropriate model design paradigm.

Despite these advances, several critical issues remain in JRSIDSR methods. A significant challenge is ignoring the difference of receptive fields [18] required for deblurring and super-resolution tasks. Indeed, blurred RS images often exhibit spatial-varying local blur, while SR tasks contend with complex global spatial degeneration. Therefore, neither Vanilla Transformer [19]-based approaches nor Convolutional Neural Network (CNN) [20]-based methods can adequately address both types of degeneration, simultaneously. Moreover, the lack of a unified theoretical analysis of the RS image degeneration process has led to chaotic model design choices for the JRSIDSR task.

Towards these limitations, we theoretically analyze the RS image degeneration process to address these concerns and develop an optimized parallel paradigm for the JRSIDSR task. In the proposed paradigm, we decompose the RS degeneration function into four key items: blur kernel, down-sampling function, correction item and image noise. Based on it, we can design the AKMD-Net as a parallel dual-branch architecture comprising four major modules, each specifically tailored for corresponding items in the proposed paradigm. The deblurring branch incorporates our carefully designed pixel-adjustable kernel block (PAKB) to estimate the spatial-varying local blur kernels. Meanwhile, in the super-resolution branch, the proposed multi-domain attention block (MDAB) obtains global contextual information enhanced with high-frequency details. By integrating PAKB and MDAB, AKMD-Net excels in both super-resolution and deblurring tasks. Extensive evaluations demonstrate that the proposed model outperforms the current state-of-the-art (SOTA) image deblurring models [21], super-resolution models [22, 23, 24] and joint models [12]. As illustrated in Fig. 1, the images reconstructed by our model exhibit superior visual performance and less error map response. To conclude, the contributions of this paper can be summarized as follows:

(1) To estimate the local and spatial-varying blur kernels in RS images, we propose the pixel-adjustable kernel block (PAKB) in the deblurring branch, which adaptively adjusts the kernel size, and therefore extracts high-quality features for removing complex blur. Specifically, the critical component of PAKB adopts a carefully designed mask-map generator (MMG), which determines the size of convolution kernels at each position.

(2) Considering the global and complex degeneration in low-resolution RS images, we introduce the multi-domain attention block (MDAB) in the super-resolution branch. This block decomposes the features into low-frequency and high-frequency components to obtain more representative representations and achieve features with global context via extra self-attention blocks.

(3) To enhance the correlative relationships in the proposed paradigm, we incorporate the adaptive feature fusion (AFF) module to facilitate the context interaction between the deblurring branch and the super-resolution branch.

(4) Finally, we wrap Wiener filter as a novel adaptive Wiener loss (AW Loss) to further enhance image quality by eliminating the noise prior in the reconstructed images.

## II. RELATED WORK

### A. Image Deblurring

Early single-image deblurring methods primarily focused on removing uniform blur. For instance, Cho et al. [25] introduced a conventional approach to overcome blind deconvolution [26], where the blur kernel and latent image were estimated from a blurred image. However, real-world image blur tends to be complex, often comprising various blur levels. Hence, Sun et al. [27] were the first to apply CNNs [28] to single-image deblurring, specifically targeting non-uniform blurring. Various methods based on CNNs [28] were developed to improve the estimation of blur kernels. Indeed, Chakrabarti [29] proposed a novel method using multi-resolution frequency decomposition to calculate the blur kernel estimation, and Lee et al. [30] enhanced the receptive field by introducing the Iterative Filter Adaptive Network (IFAN) to address spatial-varying blur and defocus blur.

Additionally, the Transformer model [19] can seize high-order contextual relationships from images, and therefore, it has been commonly used for image deblurring. For example, the Restormer [31] model utilized a multi-Dconv head transposed attention (MDTA) module to capture long-range pixel interactions. This model efficiently handles high-resolution images through multi-head self-attention layers, capturing global contextual information that requires additional time. To address this, the Uformer [32] network introduced the Locally-Enhanced Windows (LeWin) transformer module, which conducts non-overlapping window-based self-attention instead of global attention. Additionally, the Stripformer [21] model, proposed by Tsai et al., incorporated Intra-SA and Inter-SA to detect region-specific blurred patterns of varying orientations and magnitudes for handling more complex blur.

### B. Image Super-Resolution

In recent years, CNNs [28] have been widely utilized in SR tasks due to their outstanding performance. For instance, SRCNN [20], the first model to incorporate CNNs [28] in SR, has demonstrated superior results over traditional methods regarding image quality and processing speed. FSRCNN [33] improved the training speed of SRCNN [20] by introducing a deconvolution layer for up-sampling feature maps at the end of the network. However, SRCNN [20] and FSRCNN [33] is considered too shallow to extract semantic features effectively. Consequently, building upon SRCNN [20], models like VDSR [34] and RCAN [22] were developed to increase the model depth and improve receptive field coverage, thereby enhancing overall performance. RCAN [22] was the pioneering approach to integrate channel attention mechanisms for adaptive learning of channel information. To capture global information comprehensively, several significant methods with a frequency decomposition perspective have emerged in the field of SR. Indeed, Li et al. introduced the frequency-based OR-Net [35] approach, aiming to separate different frequency components and effectively compensate for information loss in low-resolution images.

However, the receptive field of the CNN [20] is limited, and thus, the Transformer [19] model has been proposed to model long-distance dependence. Recently, researchers have integrated the Transformer [19] into SR tasks and proposed many competitive methods. Chen et al. proposed the Hybrid Attention Transformer (HAT) [23] for image super-resolution, incorporating channel attention [36] and window-based self-attention [37] to capture global statistics and local details. However, RS images contain various categories and complex textures, and thus, selecting prior is critical for RS image SR. Therefore, CGC-Net [38] introduced a constrained cross-attention block to incorporate contextual information into existing deep-learning based SR methods. In particular, HSE-Net [24] leveraged single-scale and cross-scale similarities in RS images, featuring a single-scale self-similarity exploitation module (SSEM) and a cross-scale connection structure (CCS). Additionally, the FCIR [39] network focused on RS image SR through Fourier analysis, presenting a simple yet effective network based on the Transformer [19] architecture.

### C. Jointly Image Deblurring and Super-Resolution

Image deblurring and SR are essential tasks in low-level vision. Given that real-world images suffer from low resolution and blurring simultaneously, Zhang et al. introduced DB-SRN [40], a model with a feature extraction sequence and a deep dual-branch convolutional neural network designed to produce directly high-resolution images from low-resolution ones with severe blurs. Besides, ED-DSRN [14] utilized an Encoder-Decoder architecture [16] for feature extraction. However, neither of these models incorporated information fusion between the two branches in the network structure. To address this gap, Zhang et al. introduced GFN [12], which includes a gate module to fuse features from different branches. In contrast to this parallel architecture, Li et al. proposed P$^2$GAN [13], an end-to-end neural network with a serial distribution of deblurring and super-resolution branches.

### III. Proposed Method

Blurring in remote sensing images is attributed to factors intrinsic to the image acquisition process, while down-sampling function is a post-acquisition digital processing technique. Consequently, the blurring process is often considered before down-sampling [41, 42, 43]. Additionally, random noise in remote sensing images is typically statistically independent of the image pixel values [44, 45] and exhibits an additive property [46, 47]. As a result, the RS image degeneration function for the JRSIDSR task can be expressed as:

$$y = G_\downarrow(F(x)) + \varepsilon \quad (1)$$

where $x$, $F(\cdot)$, $G_\downarrow(\cdot)$ and $\varepsilon$ denote the original high-quality image, blur kernel, down-sampling function, and image noise, respectively. Generally, both blur kernel and down-sampling function are linear transformations [48, 49], hence the degeneration function can be rewritten as follows:

$$y = F(x) + G_\downarrow(x) + \widehat{FG_\downarrow}(x) + \varepsilon \quad (2)$$

where $\widehat{FG_\downarrow}(x)$ is the correlative item. The JRSIDSR task can be regarded as an ill-posed degeneration problem, with its core being learning the inverse transformation of the derived RS image degeneration function. Following this paradigm (Eq.(2)), we design AKMD-Net as a parallel dual-branch architecture comprising four major modules, each specifically tailored for different items in the proposed paradigm. Specifically, the design pattern of AKMD-Net can be articulated as follows:

$$\widehat{x} = AKMD\text{-}Net(y) \begin{cases} Debur[\underbrace{PAKB}_{\text{degree}}, \underbrace{Intra\text{-}er\ SA}_{\text{orientation}}] \stackrel{\text{for}}{=} F^{-1}(\cdot), \\ SR[\underbrace{MDAB}_{\text{spatial}}, \underbrace{RCAB}_{\text{channel}}] \stackrel{\text{for}}{=} G_\downarrow^{-1}(\cdot), \\ AFF[CBFF, FFF] \stackrel{\text{for}}{=} \widehat{FG_\downarrow^{-1}}(\cdot) \\ AWLoss \stackrel{\text{for}}{=} \min \|\widehat{\varepsilon} - \varepsilon\|_1^1 \end{cases} \quad (3)$$

AS shown in Eq.(3), each modules are specifically designed target to learn its corresponding items in our proposed degeneration function (Eq.(2)). Rather than sequential integration approach [13] where identical blocks are stacked sequentially. We utilize hybrid architectures integration approach [12] to integrate hybrid blocks into the deblurring branch and super-resolution branch, which can capture richer representation of the input data [12]. Concretely, in the deblurring branch, we integrate the proposed PAKB to capture suitable degrees of blur, as well as the Intra and Inter SA blocks [21] to parse the orientations of blur features. In the super-resolution branch, we exploit our proposed MDAB and the RCAB to extract the spatial and channel features, respectively, in order to obtain a richer representation of SR features [22]. To conclude, the primary elements of the proposed AKMD-Net can be described as: (1) A deblurring branch to estimate the inverse transformation of the blur kernel $F(\cdot)$. The primary component of this branch is the PAKB block, which adapts to different levels of local blur under varying receptive fields. (2) A super-resolution branch to learn the inverse transformation of the down-sampling function $G_\downarrow(\cdot)$. Within this branch, the multi-domain attention block (MDAB) is carefully crafted to extract global contextual information enhanced with high-frequency details. (3) An adaptive feature fusion (AFF) module is proposed to enhance the correlative features $\widehat{FG_\downarrow}(x)$ and improve the extraction of contextual relationships between the deblurring branch and the super-resolution branch. (4) We further wrap the Wiener filter [50] as an additional Wiener loss function to learn the noise $\varepsilon$.

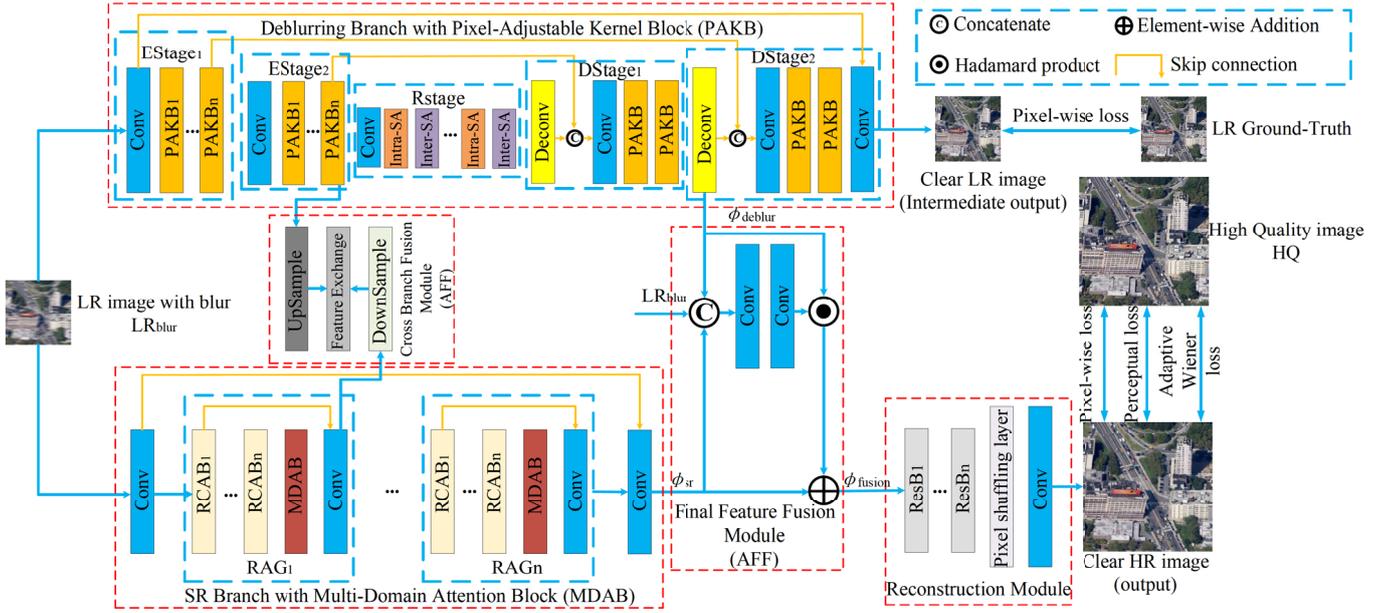

Fig. 2. The overview architecture of the proposed AKMD-Net. The features extracted by the deblurring branch $\phi_{deblur}$ and super-resolution branch $\phi_{sr}$ are fused by the final feature fusion module and then the fused feature $\phi_{fusion}$ are fed into the reconstruction module to predict the Clear HR image.

### A. Deblurring Branch with PAKB

Fig. 2 illustrates the proposed deblurring branch, which is composed of 3 stages: encoder stage (EStage), representation stage (RStage), and decoder stage (DStage), while each stage is stacked with several pixel-adjustable kernel blocks. Similar to encoder-decoder architecture, features from the encoder and decoder are concatenated via skip connections to generate multistage pixel-wise feature embeddings. In RStage, we use the Intra-SA and the Inter-SA [21] blocks to parse the orientation of the blur degeneration. For further analysis, the extracted features from the deblurring branch are denoted as $\phi_{deblur}$.

Generally, the local blur artifacts in RS images have various magnitudes. However, existing image deblurring methods, such as Stripformer [21], DeblurGAN [51], and SRN-DeblurNet [52] extract features from a standard convolution layer with fixed convolution kernel size and cannot address different magnitudes of local blur. Therefore, we develop the PAKB to handle various degrees of blur under several receptive fields. To adaptively regulate the receptive fields, we design a novel block named mask-map generator (MMG) in PAKB to produce feature mask maps with proper receptive fields. As depicted in Fig. 3, given a raw feature $X \in \mathbb{R}^{H \times W \times C}$ ($H$, $W$, $C$ are the height, width, and channel of the feature map, respectively), one convolution layer with a $1 \times 1$ kernel is firstly employed in MMG to aggregate channel information and adjust the shape of the input tensor $X \in \mathbb{R}^{H \times W \times C}$ to ($H \times W \times 1$). Then, we use a Fully Connected (FC) layer to reduce the feature blurring magnifies followed by a Layer Normalization (LN) layer for data stabilization. After that, another FC layer is appended to calculate the magnitude of blurring over pixel points. Therefore, the mask map $\widehat{M} \in \mathbb{R}^{H \times W \times 1}$ is formed as follows:

$$\widehat{M} = R\left(FC\left(FC\left(R(\delta(X))\right)\right)\right) \quad (4)$$

where $\delta$, $R$, and $FC$ represent the $1 \times 1$ convolution, reshape operation, and Fully Connected layer, respectively. All activation functions and the LN layer are omitted in Eq. (3) for brevity. To adaptively adjust the convolution kernel sizes at each position, we carefully design the Select-Kernel (SK) function with two thresholds $min_t$ and $max_t$, which classifies the magnitude of blurring into three categories and separates the three categories into three corresponding mask maps, $M_1$, $M_3$, and $M_5$ ($M_i \in \mathbb{R}^{H \times W \times 1}$ is a one-hot encoding map, representing the actuating scope of the $i \times i$ convolution kernel). And the detailed computational process of the SK function is described as lines 4-9 in Algorithm 1:

$$M_1, M_3, M_5 = \boldsymbol{SK}(\widehat{M}) \quad (5)$$

The SK function is employed in MMG to handle varying levels of blur with different kernel sizes. Fig. 3 highlights that in each PAKB, we first apply the MMG to produce the three mask maps $M_1$, $M_3$, and $M_5$. Subsequently, three convolution layers with different convolution kernel sizes are utilized to generate $X_1$, $X_3$, and $X_5$ (lines 13-15 in Algorithm 1, $X_i$ represents the feature extracted by $i \times i$ convolution kernel). Finally, to apply different convolution kernels into respective pixel positions, we multiply $X_i$ by $M_i$ to obtain $Y_i$ and add each $Y_i$ to obtain $Y$ (lines 16 in Algorithm 1):

$$Y = (X_1 \odot M_1) \oplus (X_3 \odot M_3) \oplus (X_5 \odot M_5) \quad (6)$$

where $\odot$, $\oplus$, and $Y$ are the Hadamard product, Element-wise Addition, and the final output of the proposed PAKB,

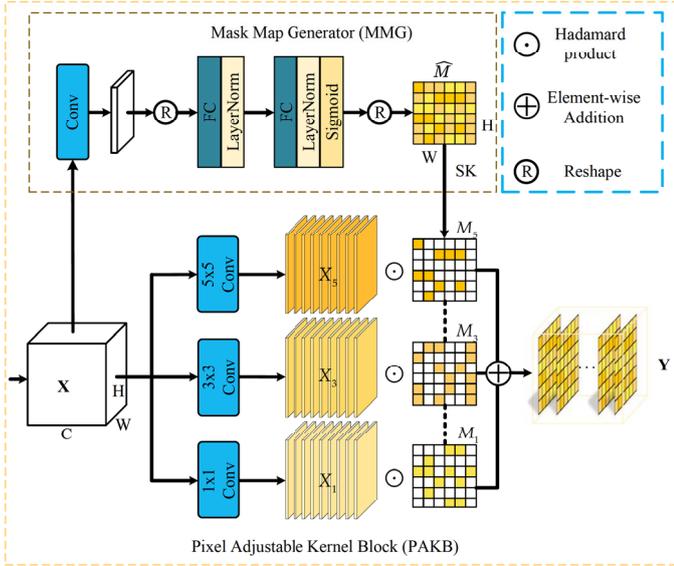

Fig. 3. Illustration of Pixel Adjustable Kernel Block (PAKB) and the workflow of Mask Map Generator (MMG).

respectively. Algorithm 1 displays the detailed description of the proposed PAKB.

---

**Algorithm 1** Pixel Adjustable Kernel Block

**Input:** raw feature $X$
1: Initialize three zero matrices $M_1, M_3, M_5$ and two thresholds $\boldsymbol{min}_t, \boldsymbol{max}_t$
2: **function** Mask-Map-Generator $(X)$:
3: $\quad \widehat{M} \leftarrow R(FC(FC(R(\delta(X)))))$
4: $\quad \forall \widehat{M}^{(i,j)} \leq \boldsymbol{min}_t$:
5: $\quad\quad M_1^{(i,j)} \leftarrow 1$
6: $\quad \forall \widehat{M}^{(i,j)} > \boldsymbol{min}_t \land \widehat{M}^{(i,j)} < \boldsymbol{max}_t$:
7: $\quad\quad M_3^{(i,j)} \leftarrow 1$
8: $\quad \forall \widehat{M}^{(i,j)} \geq \boldsymbol{max}_t$:
9: $\quad\quad M_5^{(i,j)} \leftarrow 1$
10: $\quad$ **return** $M_1, M_3, M_5$
11: **end function**
12: $M_1, M_3, M_5 \leftarrow$ Mask-Map-Generator $(X)$
13: $X_1 \leftarrow Conv_{1\times 1}(X)$
14: $X_3 \leftarrow Conv_{3\times 3}(X)$
15: $X_5 \leftarrow Conv_{5\times 5}(X)$
16: $Y \leftarrow (X_1 \odot M_1) \oplus (X_3 \odot M_3) \oplus (X_5 \odot M_5)$
17: **return** $Y$
**Output:** processed feature $Y$

---

### B. Super-Resolution Branch with MDAB

In the super-resolution branch, we first use a convolution layer to extract the shallow features from the blurry low-resolution input $LR_{blur}$. Then, a series of residual attention groups (RAGs) and a convolution layer are utilized to extract the high-dimensional features for image super-resolution. As depicted in Fig. 2, each RAG contains several residual channel attention Blocks (RCABs), one multi-domain attention block (MDAB), and a 3 × 3 convolution layer with a residual connection [53]. The extracted features from the super-resolution branch are denoted as $\phi_{sr}$.

RS images contain various types of ground objects with multi-scale details. Therefore, enhancing global contexts of low-resolution images with high-frequency details can achieve better SR results. Existing Transformer-based SR methods [23, 24] often ignore local details when capturing global contextual information. Therefore, we design the multi-domain attention block (MDAB). Due to the low-frequency content information and high-frequency detailed information are highly intertwined in RS images, we decompose the features into low-frequency components and high-frequency components via DWT. After that, we enhance the high-frequency details of RS images with global contextual information.

In detail, as shown in Fig. 4, given the input 2D feature map $X_{b-1} \in \mathbb{R}^{H \times W \times C}$, we first employ the LN layer to stabilize the data distribution in MDAB. Then, the 2D-DWT down-samples the feature map by decomposing it into four wavelet sub-bands. The content information of the feature map is stored in the low-frequency component $X_{LL} \in \mathbb{R}^{(H/2) \times (W/2) \times C}$, while detailed information is stored in three high-frequency components $X_{LH} \in \mathbb{R}^{(H/2) \times (W/2) \times C}$, $X_{HL} \in \mathbb{R}^{(H/2) \times (W/2) \times C}$, and $X_{HH} \in \mathbb{R}^{(H/2) \times (W/2) \times C}$. For the low-frequency component, we further obtain the global representation $\tilde{\chi}_{LL} \in \mathbb{C}^{(H/2) \times (W/4) \times \mathbb{C}}$ through 2D-DFT [54]:

$$\tilde{\chi}_{LL} = F[X_{LL}] \in \mathbb{C}^{(H/2) \times (W/4) \times \mathbb{C}} \quad (7)$$

where $\mathcal{F}[\cdot]$ represents the 2D-DFT [54]. Then, we modulate the spectrum with a learnable weight parameter $\lambda_c \in \mathbb{R}^{(H/2) \times (W/4) \times \mathbb{C}}$ in the frequency domain and then transform it back into the spatial domain through 2D-IDFT [55]. This process is mathematically formulated as follows:

$$G_{\text{LL}} = \mathcal{F}^{-1}[\tilde{\chi}_{\text{LL}} \odot \lambda_c] \quad (8)$$

where $\mathcal{F}^{-1}[\cdot]$ denotes the 2D-IDFT [55], and $\odot$ is the Hadamard product. For the three high-frequency components, we concatenate them together along the channel dimension, and a 5 × 5 convolution is further applied to reform $\widehat{X} \in \mathbb{R}^{(H/2) \times (W/2) \times C}$. After that, $Q_i, K_i, V_i$ are projected from $\widehat{X}$ using three 1 × 1 convolution layers:

$$\widehat{X} = \psi(concat(X_{LL}, X_{HL}, X_{HH})) \quad (9)$$

$$Q_i = \delta_i^Q(\widehat{X}), K_i = \delta_i^K(\widehat{X}), V_i = \delta_i^V(\widehat{X}) \quad (10)$$

where $\psi$ is a 5 × 5 convolution, $\delta_i$ is a 1 × 1 convolution, and $Q_i, K_i, V_i$ denote the query, key, value for $i$-th head. After that, the multi-head self-attention in the wavelet domain is performed:

$$head_i = Softmax(Q_i K_i / d) V_i \quad (11)$$

$$L_{\widehat{X}} = Multihead(Q, K, V) = \\ Linear\left(concat\left(head_1, \cdots head_h\right)\right) \quad (12)$$

where $d$ is the query dimension. Next, we concatenate the global information $G_{LL}$ and the local details $L_{\widehat{X}}$, followed by a deconvolution layer to up-sample the feature map. Finally, the layer normalization layer and Multi-Layer Perceptron (MLP) block with the residual connection [53] are employed for channel mixing. Overall, this process is formulated as:

$$X_b = MLP\left(LN\left(De(\text{concat}(G_{LL}, L_{\widehat{X}}))\right)\right) + X_{b-1} \quad (13)$$

where $De$ indicates the Deconvolution Layer, $LN$ represents the layer normalization layer, and $X_b$ is the final output of MDAB.

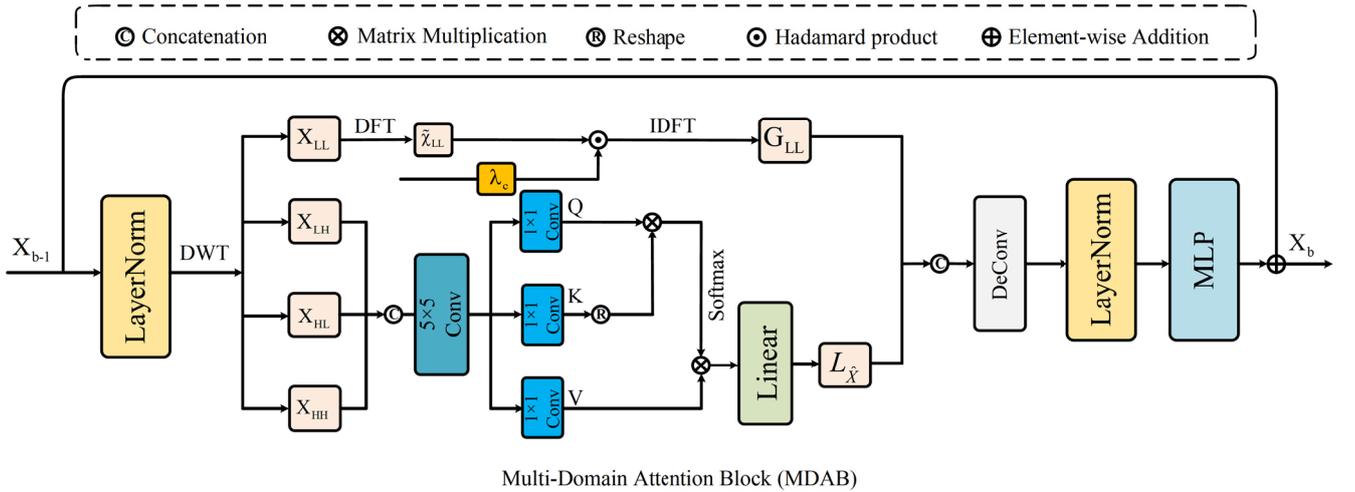

Fig. 4. The workflow of the proposed Multi-Domain Attention Block (MDAB).

### C. Adaptive Feature Fusion Module

It should be noted that Eq. (2) contains a correlative item $\widehat{FG_{\downarrow}}(x)$, which indicates the correlative degeneration in RS images. To achieve better reconstructed results, we propose an adaptive feature fusion (AFF) module to improve the contextual relationships between the deblurring branch and the super-resolution branch. As shown in Fig. 2, the AFF module is comprised of the Cross Branch Fusion Module and the Final Feature Fusion Module.

*1) Cross Branch Fusion Module:* Since the features of the two branches have different scales, the two-order feature interactions from the images are formulated as follows:

$$F_{deblur} \leftarrow \alpha * F_{deblur} + (1-\alpha) * DownSample(F_{sr}) \quad (14)$$

$$F_{sr} \leftarrow \alpha * F_{sr} + (1-\alpha) * UpSample(F_{deblur}) \quad (15)$$

where $\alpha \in [0,1]$ is a learnable parameter that adaptively exchanges contextual information between the deblurring and SR branches for high-quality contextual representation.

*2) Final Feature Fusion Module:* To investigate the characteristics of feature maps extracted from the deblurring branch and the SR branch, we visualize the feature maps of $\phi_{sr}$ and $\phi_{deblur}$ in Fig. 5. It reveals that the SR features capture spatial details from the input RS image, while the deblurring branch extracts features with more motion information. Therefore, we design an additional final feature fusion module to effectively fuse these two features. Specifically, the final feature fusion module concatenates the SR features, deblurring features, and the low-resolution (LR) blurry image as input and calculates

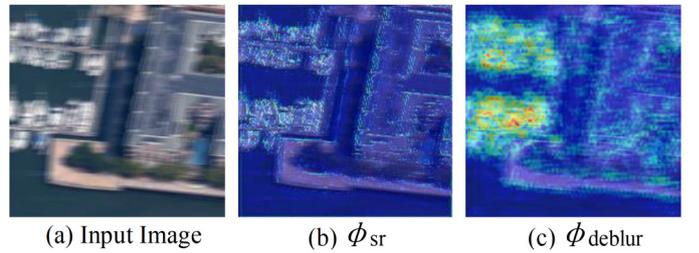

(a) Input Image  (b) $\phi_{sr}$  (c) $\phi_{deblur}$

Fig. 5. Feature response of the extracted super-resolution feature $\phi_{sr}$ and deblurring feature $\phi_{deblur}$.

the pixel-wise weight maps to adaptively select and fuse features from $\phi_{sr}$ and $\phi_{deblur}$:

$$W = \delta\left(\xi\left(concat(LR_{blur}, \phi_{sr}, \phi_{deblur})\right)\right) \quad (16)$$

$$\phi_{fusion} = W \odot \phi_{sr} \oplus \phi_{deblur} \quad (17)$$

where $\odot$ is the Hadamard product, and $\xi$ and $\delta$ denote the convolution layer with a filter size of $3 \times 3$ and $1 \times 1$, respectively.

### D. Reconstruction Module and Loss Terms

*1) Reconstruction Module:* In the reconstruction module, the fused features $\phi_{fusion}$ from the fusion module are passed through multiple Res-Blocks and pixel-shuffling layers [56] to increase the image's spatial resolution. Finally, a convolution layer reconstructs the clear HR output image.

*2) Loss Terms:* The loss functions measure the difference between the clear HR image generated from the model and the original high-quality image HQ. This paper obtains a better result with more consistency, less noise, and more semantic information by combining the total loss functions with a pixel-wise $L_1$ loss [57], a carefully designed adaptive Wiener loss (AW Loss), and a perceptual loss [58] with VGGNet19 [59].

**Pixel-wise loss.** For the JRSIDSR task, a LR image with blur $LR_{blur}$ is used as input, and the $L_1$ loss is employed to calculate the pixel-wise loss for the two tasks as follows:

$$L_{\text{pixel}} = \alpha \|G_{\theta_1}(LR_{\text{blur}}) - LR_{\text{ground -truth}}\|_1^1 \\ + \| G_{(\theta_1+\theta_2)}(LR_{\text{blur}}) - HQ \|_1^1 \quad (18)$$

where the first part is calculated on the deblurring branch, and the second part is the pixel-wise loss on the super-resolution branch. The weight parameter $\alpha$ (empirically set as 0.5) is designed to balance the two tasks. $G_{\theta_1}(LR_{\text{blur}})$ represents the output of the deblurring branch and $\theta_1$ denotes the parameter of the deblurring branch. $G_{(\theta_1+\theta_2)}(LR_{\text{blur}})$ represents the clear HR image predicted by the model and $\theta_2$ is the parameter of the super resolution branch.

**Adaptive Wiener loss.** Without a specific prior assumption of image noise, pixel-wise loss has presented limited effectiveness in reducing random noise in images [60]. Therefore, we carefully design the adaptive Wiener loss (AW Loss), which uses a Wiener filter [50] to eliminate noise $\varepsilon$ from the reconstructed images. The Wiener filter [50] is an optimal filter for random noise distributions, which depresses noise in the reconstructed images. The target denoising function of the Wiener filter [50] aims to minimize the expectation of the reconstructed error:

$$e = E\left(|F(u,v) - w(u,v) * [F(u,v) * H(u,v) + \eta(u,v)]|^2\right) \quad (19)$$

where $[F(u,v) * H(u,v) + \eta(u,v)]$ represents the degraded image in the frequecy domain, $w(u,v)$ is the denoise kernel in the frequency domain, $F(u,v)$, $H(u,v)$, and $\eta(u,v)$ are the Fourier form of the original image, degenerate filter and noise, respectively. To optimize the target function, let $\frac{\partial E(w)}{\partial w} = 0$, we can get the denoise kernel $w(u,v)$:

$$w(u,v) = \frac{H^*(u,v)}{H^2(u,v) + S_\eta/S_f} \quad (20)$$

where $H^*(u,v)$ is the complex conjugate of $H(u,v)$. $S_\eta = \eta^2(u,v)$ is the power spectrum of noise and $S_f = F^2(u,v)$ is the power spectrum of the original image. Since the power spectrum of the original image is unknown, we define the hyper-parameter $K = S_\eta/S_f$ (It is ablated in the experiment part), and $H(u,v)$ is adaptively learned by the network. With the denoise kernel $w(u,v)$, we calculate the adaptive Wiener loss (AW Loss) as follows:

$$L_{\text{aw}} = \left\| \begin{array}{c} \mathcal{F}^{-1}[\mathcal{F}[G_{(\theta_1+\theta_2)}(LR_{\text{b1ur}})]w(u,v)] \\ -\mathcal{F}^{-1}[\mathcal{F}[HQ]w(u,v)] \end{array} \right\|_1^1 \quad (21)$$

where $G_{(\theta_1+\theta_2)}(LR_{\text{b1ur}})$ represents the clear high-resolution image predicted by the proposed AKMD-Net, and HQ is the original high-quality image.

**Perceptual loss.** Pixel-wise loss [57] often leads to overly smooth [60] and poor perceptual results in reconstructing semantic details of images. Thus, to make the semantic consistency between the predicted clear HR image and the HQ image, we further use the perceptual loss [58] with a pre-trained VGGNet19 [59]:

$$L_{\text{per}} = \left\| \begin{array}{c} VGG_{(\text{i})}\left(G_{(\theta_1+\theta_2)}(LR_{\text{b1ur}})\right) \\ -VGG_{(\text{i})}(HQ) \end{array} \right\|_1^1 \quad (22)$$

where $i$ represents the $i$-th feature map of the VGG network before the activation function.

**Total loss.** With the loss functions mentioned above, we train the AKMD-Net by minimizing the joint loss functions:

$$L_{\text{total}} = L_{\text{pixel}} + \lambda_1 \cdot L_{\text{aw}} + \lambda_2 \cdot L_{\text{per}} \quad (23)$$

we empirically set the weighted coefficients $\lambda_1$ as 0.2 and $\lambda_2$ as 0.1 to represent the contributions of each loss function.

## IV. EXPERIMENT AND ANALYSIS

### A. Experimental Setup

*1) Datasets:* To validate the performance of the proposed AKMD-Net model, we utilize two commonly-used remote sensing image datasets, RSSCN7 [61] and WHU Building datasets [62]. The RSSCN7 dataset comprises a total number of 2,800 images with a fixed resolution of $400 \times 400$. From this dataset, 700 images are randomly selected for testing and validating, while the rest are used for training. LR images with scale factors of 2 and 4 are obtained by applying the Bicubic [49] function to down-sample the original HQ images. Additionally, $LR_{blur}$ images are created from LR images via random motion blur filter kernels. The WHU Building dataset has 8,189 images that extract over 220,000 buildings from New Zealand. The dataset is divided into a training set (4,736 tiles), a test set (2,416 tiles) and a validation set (1,037 tiles) following the official guidelines. In contrast to the RSSCN7 dataset [61], which contains a limited number of images, the image samples in the large-scale WHU Building dataset [62] are more likely to exhibit more complex real-world degeneration. As the most common blur in RS images, atmospheric turbulence blur [8] typically arises from variations in temperature, pressure, and wind speed within the atmosphere. Therefore, to simulate real-world remote sensing image degeneration, we utilize the method proposed by Mao et al. [63] to introduce extra atmospheric turbulence blur (with fried parameter $R_0 = 0.125$) into the WHU Building dataset. Hence the degeneration images in the WHU Building dataset contain random degeneration of motion blur, atmospheric turbulence blur, and Bicubic down-sample. Representative image samples are illustrated in Fig. 6.

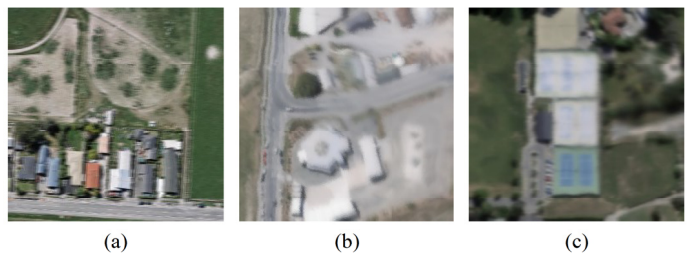

Fig. 6. (a) the low-resolution motion blur image. (b) the low-resolution atmospheric turbulence blur image (c) the low-resolution hybrid blur image (motion blur and atmospheric turbulence blur).

*2) Implementation Details and Metrics:* Generally, current SR methods and deblurring methods mainly follow the same encoder-decoder framework, hence we compare the proposed AKMD-Net with various methods, which are optimized for super-resolution (RCAN [22], HAT [23], and HSENet [24]), designed for image deblurring method (Stripformer [21]), and the jointly image deblurring and super-resolution method (GFN [12]). All models are retrained on the same training dataset with identical configurations for fair comparisons. All models are trained on the Ubuntu 22.04 system with two NVIDIA GTX 3090 GPUs. The Adam optimizer is used with $\beta_1 = 0.9$ and $\beta_2 = 0.99$ for model optimization, and the number of iterations is set to 400,000. The learning rate is initialized to $8 \times 10^{-5}$ and iteratively decreases with the MultiStepLR learning rate adjustment schedule. To quantitatively analyze the model performance, we utilize PSNR, SSIM, and RMSE to evaluate the pixel variance between the reconstructed and original high-quality images. Following some recent works [64], we also employ LPIPS [65] and VIF [66] to assess the perceptual reconstruction ability of each model.

### B. Comparison with State-of-the-Art Methods

*1) SOTA Comparison on the RSSCN7 Dataset:* Table I evaluates the performance of the proposed AKMD-Net and other comparable methods on the RSSCN7 Dataset. The Transformer-based super-resolution method HAT achieves a PSNR score of 25.3359 dB, notably 5.5367 dB higher than the basic Bicubic method at a scale factor of 2. Similarly, the PSNR score (25.8604dB) of the Transformer-based deblurring method Stripformer [21] is almost 31% higher than the Bicubic [49] method (19.7992dB) at a scale factor of 2. However, due to the complex textures and details of RS images, even the current SOTA method HSENet [24], which is explicitly designed for RS images cannot gain satisfactory results on either $2\times$ or $4\times$ scales. Unsurprisingly, CNN-based methods like GFN [12] do not work well on the JRSIDSR tasks either, which only reports a lower PSNR score of 24.3188 dB on the RSSCN7 Dataset at the scale factor of 2. Rather than the methods above, the RCAN [22] model achieves a sub-optimal PSNR score of 26.8382 dB at $\times 2$ JRSIDSR tasks by incorporating the channel attention mechanism [36]. When we test the performance of hybrid methods, which cascades one SR method and one deblurring method, it is observed that the SR-first strategy generally outperforms the deblurring-first strategy. This intriguing experimental phenomenon primarily hinges on two key factors. On the one hand, the key of the JRSIDSR task relies on learning the inverse transformation of image degeneration function. And the SR-first strategy follows the inverse RS images degeneration process of blurring before down-sampling. On the other hand, the super-resolution operation can generate local details in images that is beneficial to recover the blurred features. However, the deblurring operation can only restore information lost due to blurring, which offers limited advantages to the super-resolution process. However, it is important to note that the SR-first strategy incurs higher computational costs due to the larger feature map in the deblurring process. Compared with other methods, the proposed AKMD-Net method achieves new state-of-the-art (SOTA) results at $\times 2$ and $\times 4$ tasks on all 5 metrics (PSNR, SSIM, LPIPS, RMSE, VIF). At the same time, the reconstructed images of each model visualized in Fig. 7 demonstrate that AKMD-Net exhibits better visual quality with clearer textures and boundaries.

*2) SOTA Comparison on the WHU Building Dataset:* To further validate the effectiveness of the proposed AKMD-Net model, we evaluate the performances of these methods on the more challenging WHU Building Dataset. The results of all models are presented in Table II, showing that all models struggle with this dataset, while the proposed AKMD-Net still stands out with magnificent performance. In the $2\times$ task, AKMD-Net outperforms HSENet [24] + Stripformer [21] by 0.483 dB in PSNR score. As to the $4\times$ task, the performance of the AKMD-Net also shows consistent performance improvement across all evaluation metrics, which strongly proves the effectiveness of AKMD-Net. Furthermore, we visualize the reconstructed high-quality images of the WHU Building Dataset at $2\times$ on Fig. 8, where error maps of AKMD-Net and other state-of-the-art methods are displayed in the last row. The visualized results demonstrate that AKMD-Net performs best in texture areas and has minimal reconstruction errors.

*3) Generalizability in real-world scenarios:* To further demonstrate the progressiveness of the proposed AKMD-Net, we study the adaptability of the proposed method to multiple complex degradation conditions, including rainy environments and different noise levels. For this task, we use the Rain200H [10] dataset, which consists of 1800 training images and 200 testing images. Additionally, the training and testing data is further processed by adding Gaussian noise with standard deviation $\sigma = 10, 30, 50$. We compare the AKMD-Net performance with several competing methods [70, 21, 69]. As reported in Table III, the proposed AKMD-Net achieves consistent SOTA performance under various noise levels, which strongly demonstrate that the proposed method has better robustness and generalization ability when dealing with multiple complex degeneration conditions.

Also, to evaluate the generalizability of the proposed AKMD-Net in practical applications, we further visualize the performance of the JRSIDSR methods in real-world scenarios. Specifically, we present the reconstruction results on our recently captured RS images, which do not have corresponding ground truth image. Fig. 9 reveals that the real-world RS images reconstructed by HSENet [24]+Stripformer [21] achieves clearer textures than RCAN [22] and GFN model [12]. However, it can be obviously observed that the proposed AKMD-Net has the best visual effect on reconstructing the texture information among them, which provides its generalization ability and potential to reconstruct diverse degraded RS images in real-world scenarios.

### C. Ablation Study

In this study, we analyze the significance of each component in the proposed AKMD-Net by conducting ablation experiments on the RSSCN7 Dataset with a scale factor of 4.

*1) Component Ablations:* To clarify the individual contributions of each component and justify their inclusion in the model., we conduct component ablation experiments for PAKB, MDAB, AFF (Cross-Branch Feature Fusion (CBFF) Module and Final Feature Fusion (FFF) Module), and the AW

TABLE I
QUANTITATIVE COMPARISON OF STATE-OF-THE-ART METHODS ON THE RSSCN7 DATASET. THE BEST RESULTS ARE HIGHLIGHTED IN RED, WHILE THE SUBOPTIMAL RESULTS ARE MARKED IN BLUE

| Method | Scale | PSNR↑ | SSIM↑ | LPIPS↓ | RMSE↓ | VIF↑ |
|---|---|---|---|---|---|---|
| Bicubic [49] (SR) | ×2 | 19.7992 | 0.5564 | 0.6778 | 0.1269 | 0.7957 |
| RCAN [22] (SR) | ×2 | 26.8382 | 0.7745 | 0.3385 | 0.0585 | 0.8216 |
| HAT [23] (SR) | ×2 | 25.3359 | 0.7117 | 0.3934 | 0.0690 | 0.8117 |
| HSENet [24] (SR) | ×2 | 25.1792 | 0.7135 | 0.3856 | 0.0703 | 0.812 |
| Stripformer [21] (Deblur) | ×2 | 25.8604 | 0.7466 | 0.3436 | 0.0655 | 0.8121 |
| GFN [12] (Joint) | ×2 | 24.3188 | 0.713 | 0.3832 | 0.076 | 0.8079 |
| Stripformer [21] (Deblur) + HSENet [24] (SR) | ×2 | 25.8261 | 0.7148 | 0.3898 | 0.0697 | 0.8113 |
| HSENet [24] (SR) + Stripformer [21] (Deblur) | ×2 | 26.6953 | 0.7775 | 0.3321 | 0.0592 | 0.8186 |
| RealSR [67] (RWISR) | ×2 | 26.5874 | 0.7732 | 0.3371 | 0.0693 | 0.8162 |
| SRResCGAN [68] (RWISR) | ×2 | 26.4434 | 0.7728 | 0.3389 | 0.0723 | 0.8121 |
| Shoyan, Mekhak, et al. [69] (Joint) | ×2 | 26.2245 | 0.7699 | 0.3413 | 0.0788 | 0.8098 |
| Ours | ×2 | **27.6447** | **0.7978** | **0.3207** | **0.0521** | **0.8280** |
| Bicubic [49] (SR) | ×4 | 18.0951 | 0.4756 | 0.8191 | 0.1510 | 0.7909 |
| RCAN [22] (SR) | ×4 | 23.5869 | 0.6031 | 0.5334 | 0.0843 | 0.8025 |
| HAT [23] (SR) | ×4 | 22.8901 | 0.5780 | 0.5846 | 0.0917 | 0.8005 |
| HSENet [24] (SR) | ×4 | 22.9281 | 0.5793 | 0.5706 | 0.0912 | 0.7992 |
| Stripformer [21] (Deblur) | ×4 | 23.7321 | 0.6013 | 0.5246 | 0.0840 | 0.8049 |
| GFN [12] (Joint) | ×4 | 22.5106 | 0.5804 | 0.5457 | 0.0972 | 0.8004 |
| Stripformer [21] (Deblur) + HSENet [24] (SR) | ×4 | 22.8829 | 0.5777 | 0.5668 | 0.0919 | 0.8014 |
| HSENet [24] (SR) + Stripformer [21] (Deblur) | ×4 | 23.1798 | 0.5912 | 0.5295 | 0.0895 | 0.8021 |
| RealSR [67] (RWISR) | ×4 | 23.0453 | 0.5892 | 0.5325 | 0.0907 | 0.8014 |
| SRResCGAN [68] (RWISR) | ×4 | 22.9122 | 0.5834 | 0.5333 | 0.0915 | 0.8001 |
| Shoyan, Mekhak, et al. [69] (Joint) | ×4 | 22.7148 | 0.5815 | 0.5235 | 0.0995 | 0.7981 |
| Ours | ×4 | **24.5210** | **0.6278** | **0.5133** | **0.0769** | **0.8116** |

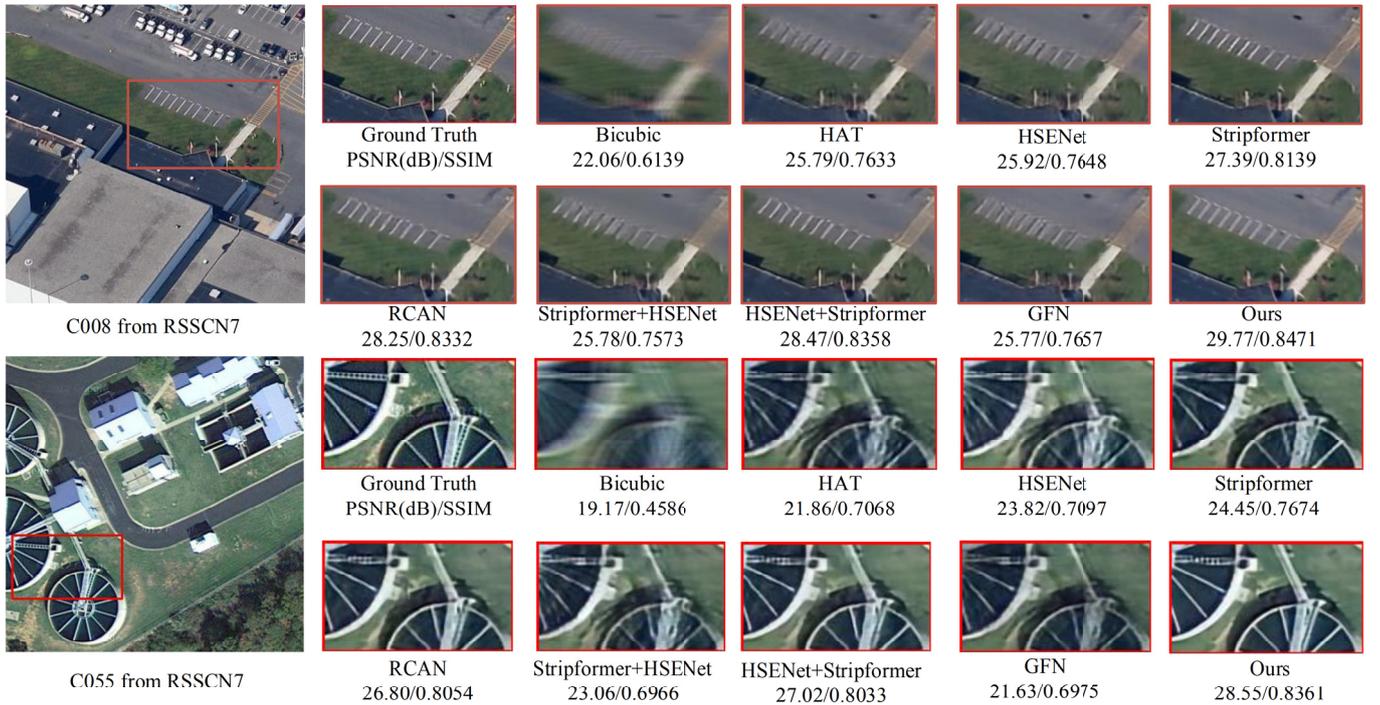

Fig. 7. Visual comparison on the RSSCN7 Dataset with a scale factor of 2. The patches for comparison are marked with red boxes in the original images. PSNR/SSIM values are calculated based on the patches to more accurately reflect the performance difference.

Loss function in the proposed AKMD-Net. We incrementally introduce these modules and conduct six experiments, with the results presented in Table IV. Initially, we replaced the original convolution layer in the deblurring branch of the baseline model [21] with PAKB, resulting in a 0.2783 dB PSNR improvement. This improvement is attributed to the ability of the proposed PAKB to handle different magnitudes of local blur that require varying receptive field sizes. Furthermore, we visualize the mask maps from MMG and the feature map after performing PAKB in Fig. 10, where larger values in $\widehat{M}$ correspond to more severe blur areas. Meanwhile, the activation map after PAKB illustrates a higher response in regions with larger blur. Then, we can see from Table IV that MDAB in the SR branch leads to a 0.3418 dB PSNR increment, which demonstrates the necessity of obtaining global contextual information enhanced with high-frequency details through multi-domain (spatial domain, frequency domain, and wavelet domain) learning. In addition, the CBFF module in AFF facilitates feature interaction between the dual branches and achieves a higher PSNR of 24.1775 dB. At the same time, when use FFF module to fuse SR features $\phi_{SR}$ and deblur features $\phi_{deblur}$ rather than simply add them, the PSNR score be can observed an additional 0.0912 dB bonus. Finally, applying the AW Loss obtains an extra 0.9957 dB PSNR improvement over the baseline. Besides, Fig. 11 illustrates the PSNR comparisons of the AKMD-Net with the different proposed components throughout the training phase, which emphasizes the essential nature of all components for the proposed AKMD-Net.

TABLE II
Quantitative comparison of state-of-the-art methods on the WHU Building dataset. The best results are highlighted in red, while the suboptimal results are marked in blue

| Method | Scale | PSNR↑ | SSIM↑ | LPIPS↓ | RMSE↓ | VIF↑ |
|---|---|---|---|---|---|---|
| Bicubic [49] (SR) | × 2 | 20.3025 | 0.4119 | 0.7641 | 0.1179 | 0.8440 |
| RCAN [22] (SR) | × 2 | 22.5405 | 0.5543 | 0.4800 | 0.0929 | 0.8771 |
| HAT [23] (SR) | × 2 | 22.0665 | 0.5239 | 0.5149 | 0.0981 | 0.8668 |
| HSENet [24] (SR) | × 2 | 22.5186 | 0.5417 | 0.4982 | 0.0940 | 0.8754 |
| Stripformer [21] (Deblur) | × 2 | 22.0149 | 0.5243 | 0.5050 | 0.0985 | 0.8701 |
| GFN [12] (Joint) | × 2 | 22.0594 | 0.5285 | 0.5160 | 0.0979 | 0.8640 |
| Stripformer [21] (Deblur) + HSENet [24] (SR) | × 2 | 22.3750 | 0.5377 | 0.5066 | 0.0954 | 0.8760 |
| HSENet [24] (SR) + Stripformer [21] (Deblur) | × 2 | 22.5565 | 0.5491 | 0.4868 | 0.0937 | 0.8756 |
| RealSR [67] (RWISR) | × 2 | 22.4019 | 0.5487 | 0.4877 | 0.0944 | 0.8745 |
| SRResCGAN [68] (RWISR) | × 2 | 22.2798 | 0.5472 | 0.4895 | 0.0957 | 0.8739 |
| Shoyan, Mekhak, et al. [69] (Joint) | × 2 | 22.1998 | 0.5463 | 0.5032 | 0.0946 | 0.8731 |
| Ours | × 2 | **23.0395** | **0.5771** | **0.4504** | **0.0911** | **0.8836** |
| Bicubic [49] (SR) | × 4 | 18.7385 | 0.3138 | 0.9663 | 0.1417 | 0.8290 |
| RCAN [22] (SR) | × 4 | 20.4035 | 0.4057 | 0.6992 | 0.1179 | 0.8503 |
| HAT [23] (SR) | × 4 | 20.1296 | 0.3781 | 0.7522 | 0.1215 | 0.8451 |
| HSENet [24] (SR) | × 4 | 20.2667 | 0.3885 | 0.7320 | 0.1197 | 0.8480 |
| Stripformer [21] (Deblur) | × 4 | 20.1568 | 0.3778 | 0.7780 | 0.1209 | 0.8450 |
| GFN [12] (Joint) | × 4 | 20.1882 | 0.3859 | 0.7279 | 0.1207 | 0.8449 |
| Stripformer [21] (Deblur) + HSENet [24] (SR) | × 4 | 20.3592 | 0.3921 | 0.7253 | 0.1185 | 0.8500 |
| HSENet [24] (SR) + Stripformer [21] (Deblur) | × 4 | 20.5819 | 0.4065 | 0.6921 | 0.1163 | 0.8531 |
| RealSR [67] (RWISR) | × 4 | 20.4738 | 0.4053 | 0.6943 | 0.1170 | 0.8527 |
| SRResCGAN [68] (RWISR) | × 4 | 20.3409 | 0.4052 | 0.6948 | 0.1183 | 0.8519 |
| Shoyan, Mekhak, et al. [69] (Joint) | × 4 | 20.3122 | 0.4037 | 0.6954 | 0.1197 | 0.8513 |
| Ours | × 4 | **20.9569** | **0.4294** | **0.6630** | **0.1113** | **0.8560** |

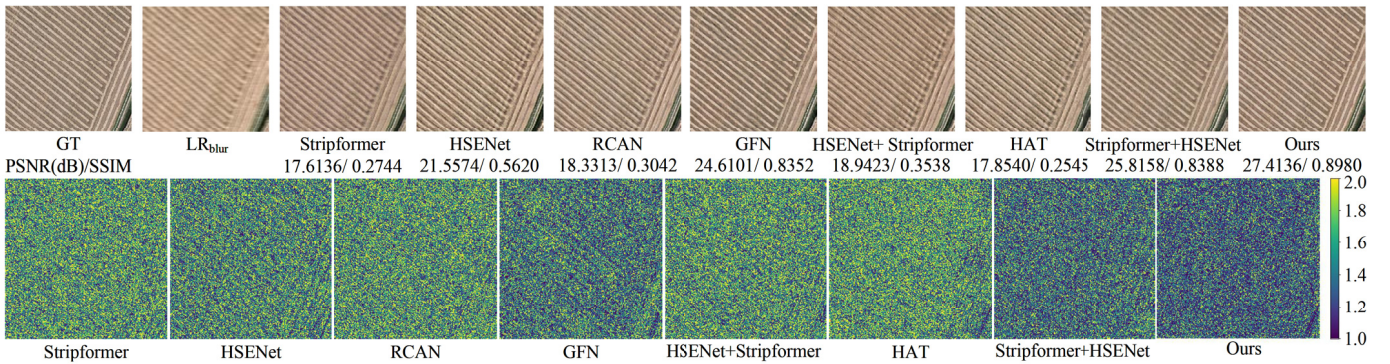

Fig. 8. The visualization results and error maps of different methods on the WHU Building Dataset. Ground Truth image, Blurry LR input, and the reconstructed images are illustrated on the first rows. The error heat maps are located in the last row.

*2) Choice of Kernel Size:* To further investigate the impact of different kernel sizes on PAKB performance, we conduct experiments with various combinations of convolution kernel sizes in the PAKB module. The results of these experiments are presented in Table V. Among the different combinations, the group using kernel sizes of 3, 5, and 7 achieves the highest PSNR of 24.71dB and SSIM value of 0.69 on the RSSCN7 dataset, which outperforms our initial configuration. Conversely, the initial group, which utilizes kernel sizes of 1, 3, and 5, yields unsatisfactory results possibly due to the small receptive field of the convolution kernel sizes ($1 \times 1$), which limits the extraction of contextual information for deblurring features. On the other hand, the third group, which includes overlarge convolution kernel size ($9 \times 9$), may have extracted excessive noise during feature extraction and leads to a reduction in model performance.

TABLE III
AVERAGE PSNR AND SSIM RESULTS OF $\sigma$ 10, 30, 50 ON RAIN200H DATASET.

| Method | Rain200H dataset | | |
|---|---|---|---|
| | $\sigma=10$ | $\sigma=30$ | $\sigma=50$ |
| GFN [70] | 30.95/0.9120 | 27.89/0.7344 | 26.74/0.6633 |
| Stripformer [21] | 32.00/0.9345 | 29.78/0.7569 | 28.01/0.6801 |
| HSENet [24]+Stripformer [21] | 32.27/0.9389 | 29.91/0.7587 | 28.87/0.6894 |
| Stripformer [21]+HSENet [24] | 31.90/0.9331 | 29.79/0.7566 | 28.09/0.6804 |
| Shoyan, Mekhak, et al [69] | 32.22/0.9374 | 29.83/0.7574 | 28.21/0.6833 |
| Ours | **32.87/0.9430** | **30.69/0.8219** | **29.66/0.7030** |

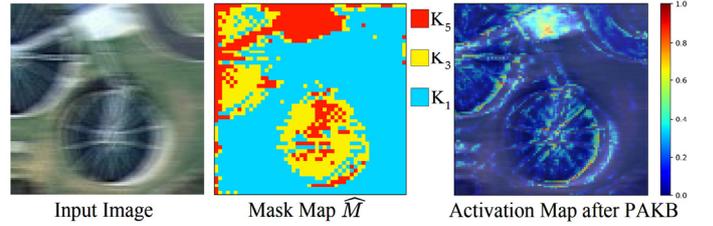

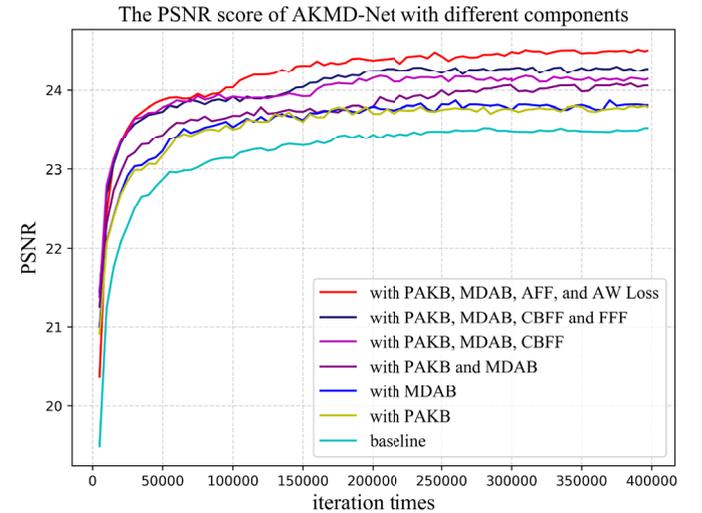

Fig. 10. The result of PAKB's adaptation to different degrees of blur. $K_1$, $K_3$, $K_5$ represent the convolution kernel sizes with $1 \times 1$, $3 \times 3$, and $5 \times 5$, respectively.

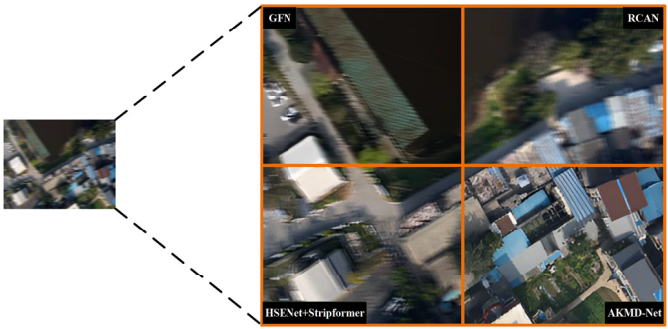

Fig. 9. Visualization of JRSIDSR results on images in the real scene. All the JRSIDSR models are pre-trained on the WHU Building dataset.

Fig. 11. The PSNR score tendency of AKMD-Net with different components during the training period.

*3) Type of Wavelet Families:* In this section, we investigate the impact of different types of wavelet families on the image reconstruction results. Various wavelet families are available for 2D-DWT, with our investigation focusing on three common ones: Haar, Biorthogonal, and Daubechies. Specifically, "Haar" denotes the simplest Haar wavelet, which consists of two rectangle functions. "biorNr.Nd" denotes the Biorthogonal wavelet with orders (Nr, Nd). The filter coefficients of Biorthogonal wavelet functions are more complex than other wavelet functions and thus have higher computational complexity. And "dbN" is the Daubechies wavelet with approximation order N. The experimental results are summarized in Table VI, showing that the Daubechies wavelet (db3) achieves the highest PSNR and SSIM scores. It improves the PSNR and SSIM scores of the Haar wavelet by 0.4% and 0.35%, respectively. Generally, the Daubechies wavelet with a higher order N exhibits enhanced localization in the frequency field, which results in improved frequency band division and superior experimental outcomes.

*4) Setting of Hyperparameter K:* From the standard Wiener filter [50] formula (Eq. 19), K represents the power ratio of

TABLE IV
COMPONENTS ABLATION RESULTS OF AKMD-NET.

| Baseline | PAKB | MDAB | AFF | | AW Loss | PSNR↑ |
|---|---|---|---|---|---|---|
| | | | CBFF | FFF | | |
| ✓ | ✗ | ✗ | ✗ | ✗ | ✗ | 23.5253 |
| ✓ | ✓ | ✗ | ✗ | ✗ | ✗ | 23.8036 |
| ✓ | ✗ | ✓ | ✗ | ✗ | ✗ | 23.8734 |
| ✓ | ✓ | ✓ | ✗ | ✗ | ✗ | 24.0854 |
| ✓ | ✓ | ✓ | ✓ | ✗ | ✗ | 24.1775 |
| ✓ | ✓ | ✓ | ✓ | ✓ | ✗ | 24.3053 |
| ✓ | ✓ | ✓ | ✓ | ✓ | ✓ | **24.5210** |

TABLE V
THE RESULT OF AKMD-NET WITH DIFFERENT COMBINATIONS OF MULTIPLE KERNELS. K1, K3, K5, AND K7 REPRESENT THE KERNEL SIZES OF 1, 3, 5, 7, RESPECTIVELY.

| K1 | K3 | K5 | K7 | K9 | PSNR↑ | SSIM↑ |
|---|---|---|---|---|---|---|
| ✓ | ✓ | ✓ | | | 24.52 | 0.63 |
| | ✓ | ✓ | ✓ | | **24.71** | **0.69** |
| | | ✓ | ✓ | ✓ | 24.48 | 0.62 |

TABLE VI
PERFORMANCE COMPARISON OF THE MDAB USING DIFFERENT WAVELETS.

| Wavelet Type | | PSNR↑ | SSIM↑ |
|---|---|---|---|
| Haar | | 24.5210 | 0.6278 |
| Biorthogonal | bior2.2 | 24.4978 | 0.6293 |
| | bior3.3 | 24.4201 | 0.6297 |
| Daubechies | db2 | 24.5153 | 0.6284 |
| | db3 | **24.6215** | **0.6300** |

TABLE VII
ABLATION STUDY ON DIFFERENT HYPERPARAMETER SETTINGS OF THE PROPOSED AW LOSS.

| K= $S_\eta/S_f$ | PSNR↑ | SSIM↑ |
|---|---|---|
| K = 0.08 | 24.3447 | 0.6240 |
| K = 0.09 | 24.5071 | 0.6275 |
| K = 0.10 | **24.5210** | **0.6278** |
| K = 0.11 | 24.4431 | 0.6275 |
| K = 0.12 | 24.4955 | 0.6272 |

TABLE VIII
EFFECTIVENESS OF THE AW LOSS. W/O MEANS WITHOUT AND W/ MEANS WITH.

| Method | PSNR↑ | SSIM↑ | LPIPS↓ |
|---|---|---|---|
| Deblurring Brach (w/o AW Loss) | 23.9170 | 0.6047 | 0.5217 |
| Deblurring Brach (w/ AW Loss) | 24.0045 | 0.6055 | 0.5201 |
| SR Brach (w/o AW Loss) | 23.7332 | 0.6027 | 0.5233 |
| SR Brach (w/ AW Loss) | 23.9018 | 0.6045 | 0.5221 |
| Jointly Dual-Branch (w/o AW Loss) | 24.3053 | 0.6234 | 0.5155 |
| Jointly Dual-Branch (w/ AW Loss) | **24.5210** | **0.6278** | **0.5133** |

noise to signal at each point of the image, i.e., the Fourier spectrum of the reciprocal of the signal-to-noise ratio. In theory, the K value should vary for each image, but practically, individualizing it would significantly increase algorithm complexity and system resource consumption. Hence, our study focuses on estimating the average signal-to-noise ratio across all dataset images. Table VII illustrates that AKMD-Net achieves the highest PSNR (24.5210dB) and SSIM (0.6278) values when K is set at 0.10. Decreasing K to 0.08 results in subpar PSNR (24.3447dB) and SSIM (0.6240) metrics compared to the optimal values. Similarly, setting the K value of 0.09 yields inferior results compared to when K is 0.10, as the K value influences the frequency response of the Wiener filter. Furthermore, increasing K moderately to 0.11 and 0.12 leads to decreased performance of AKMD-Net. We believe it is because the larger K value enhances noise suppression in the image, which is an obstacle to preserving image details. To conclude, setting the K value of 0.10 gains a great trade off between noise suppression and image details preservation.

*5) Effectiveness of the AW Loss:* The AW Loss aims to depress the prior noise in the reconstructed images, making it more likely to render high-quality reconstruction results on the JRSIDSR task. Here, we explore whether the AW Loss will contribute to the performance metrics. To this end, we study the effect of the AW Loss on deblurring branch, super-resolution branch and the overall joint model, respectively. The quantitative results can be found in Table VIII. As we can see, only use one deblurring branch or super-resolution branch to tackle the JRSIDR task cannot get satisfied results compared to the Jointly Dual-Branch model (AKMD-Net). This is reasonable, as single branch struggles to deal with the different demands of receptive-fields between the deblurring task and the super-resolution task. Meanwhile, it can be observed that all different methods takes more improvements in terms of PSNR, SSIM and LPIPS by ultizing the proposed AW Loss to train these methods, which strongly demonstrates the potential of integrating the AW Loss into low-level tasks.

*6) Adaptability to different scales:* Due to the varying resolutions of RS images, we investigate the adaptability of the proposed AKMD-Net across different input scales. We compare the proposed AKMD-Net with several competing methods [21, 24, 69] on three different input resolution conditions: 48×48, 96×96, and 128×128. Experimental results in Table IX show that our proposed method achieves consistent performance advantages on all input scales, which suggests AKMD-Net's robustness and adaptability to reconstruct different scales of RS images.

### D. Limitations and Disscussions

*1) Model Efficiency Analysis:* To assess the computational complexity of the AKMD-Net, we present the parameters and floating-point operations per second (FLOPs) in Table X. The CNN-based joint method GFN has the lowest parameter count and FLOPs. While the HSENet-first strategy generally outperforms the Stripformer-first strategy in image reconstruction, it is worth noting that the former incurs higher FLOPs due to image deblurring in high-resolution space. Our proposed AKMD-Net strikes a great trade-off between model performance and computational cost. It offers excellent reconstruction results without introducing excessive computational burden. Furthermore, to evaluate the model's efficiency in large-scale RS applications, we increase the image resolution and compare the FLOPs of the GFN, Stripformer + HSENet, and our proposed AKMD-Net. The results in Fig. 12 indicate that the computational complexity of our proposed AKMD-Net shows a significant disadvantage as the input resolution increases. While the proposed MDAB enhances image reconstruction performance, the implementation of various transformations, such as wavelet and Fourier transforms, introduces an additional computational burden. Theoretically, the computational complexity of DWT, FFT, and IFFT is $O(n^2)$, $O(nlog(n))$ and $O(nlog(n))$, respectively. Performing self-attention in our proposed MDAB involves the computation complexity of $[(O((1/4)n)^2) = (1/16)O(n^2)]$. Thus, the total computational complexity of our proposed MDAB is about $2nlogn + (17/16)O(n^2)$, which is slightly higher than the naive SA $O(n^2)$, and $n$ is the size of patches. Therefore, improving the computational efficiency is one of the most critical areas for our future work of the proposed AKMD-Net.

TABLE IX
THE ADAPTABILITY OF THE PROPOSED AKMD-NET TO DIFFERENT INPUT SCALES.

| Input Scale | Methods | PSNR↑ | SSIM↑ |
|---|---|---|---|
| 48×48 | Stripformer [21] + HSENet [24] | 20.37 | 0.5074 |
| | HSENet [24] + Stripformer [21] | 20.78 | 0.5123 |
| | Shoyan, Mekhak, et al. [69] | 20.51 | 0.5090 |
| | Ours | **21.34** | **0.5207** |
| 96×96 | Stripformer [21] + HSENet [24] | 21.20 | 0.5111 |
| | HSENet [24] + Stripformer [21] | 21.73 | 0.5289 |
| | Shoyan, Mekhak, et al. [69] | 21.32 | 0.5170 |
| | Ours | **22.21** | **0.5335** |
| 128×128 | Stripformer [21] + HSENet [24] | 22.32 | 0.5789 |
| | HSENet [24] + Stripformer [21] | 22.77 | 0.5821 |
| | Shoyan, Mekhak, et al. [69] | 22.52 | 0.5801 |
| | Ours | **22.97** | **0.5923** |

*2) Disscussions of potential failure cases:* While the proposed AKMD-Net exhibits new SOTA results on the JR-SIDSR task compared with many existing methods. It has two potential limitations related to the degeneration function presented in Eq. (1). First, the RS image degeneration process is formulated in Eq. (1) based on the fundamental assumption that the noise $\varepsilon$ in RS images is additive. This implies that AKMD-Net may struggle to reconstruct RS images that contain multiplicative noise [71], such as speckle noise [72] and rayleigh noise [73]. In addition, to simplify the degeneration process form $x$ to $y$ in Eq. (1), we regard $F(\cdot)$ as a simple linear blur degeneration process within the context of the deblurring task. Consequently, more complex degeneration functions, such as those associated with low-light conditions [74], and hazy scenes [75], may hinder the reconstruction performance of the proposed AKMD-Net.

### E. Effects on RS downstream tasks

In this section, to demonstrate that high-quality RS images have practical benefits for achieving superior results in real-world RS tasks, we conduct experiments on the RS Scene Classification task and RS Semantic Segmentation task.

TABLE X
PARAMETER AND FLOPS ANALYSIS. FLOPS ARE TESTED ON A LR IMAGE WITH 48 × 48 PIXELS.

| Method | Scale | Param (M) | FLOPs (G) |
|---|---|---|---|
| GFN [12] | ×2 | 12.06 | 8.02 |
| Stripformer [21] + HSENet [24] | ×2 | 24.99 | 15.36 |
| HSENet [24] + Stripformer [21] | ×2 | 24.99 | 33.29 |
| Ours | ×2 | 26.09 | 19.29 |
| GFN [12] | ×4 | 12.21 | 10.45 |
| Stripformer [21] + HSENet [24] | ×4 | 25.14 | 16.77 |
| HSENet [24] + Stripformer [21] | ×4 | 25.14 | 106.41 |
| Ours | ×4 | 26.24 | 21.72 |

*1) Effects on Remote Sensing Scene Classification:* Low-level vision tasks are typically seen as an essential preprocessing step for high-level vision tasks, such as semantic segmentation [1], object detection [2], and image classification [3]. Specifically, when dealing with low-quality images as inputs, low-level methods can enhance image details, thus benefiting high-level tasks. To validate the effectiveness of our proposed AKMD-Net method on high-level vision tasks, we conduct experiments on the RS scene classification task within the NWPU-RESISC45 (NWPU) dataset, which comprises 45 categories with 700 images per class and the resolution of every image is a fixed 256×256. In this experiment, we randomly divide the NWPU dataset into training and testing sets. Then we finetune the ResNet-50 [53] model on the training set, which has been pretrained on ImageNet. During the test phase, we take the original images as HQ images, and the LQ images are produced by Bicubic downsampling function with scale factor of 2× and random motion blur filter kernels. The LQ images are then reconstructed via multiple methods, including bicubic interpolation [49], RCAN [22], HAT [23], HSENet [24], Stripformer [21], GFN [12], Stripformer [21] + HSENet [24], HSENet [24] + Stripformer [21], and our proposed AKMD-Net. Then, we use the fine-tuned ResNet-50 model as a classifier for the reconstruction RS images and the classification results are summarized in Table XI. It can be observed that the ResNet-50 model achieves the best classification accuracy when testing on the images generated by the proposed AKMD-Net, which suggests that the proposed AKMD-Net can significantly enhances the quality of the degraded images.

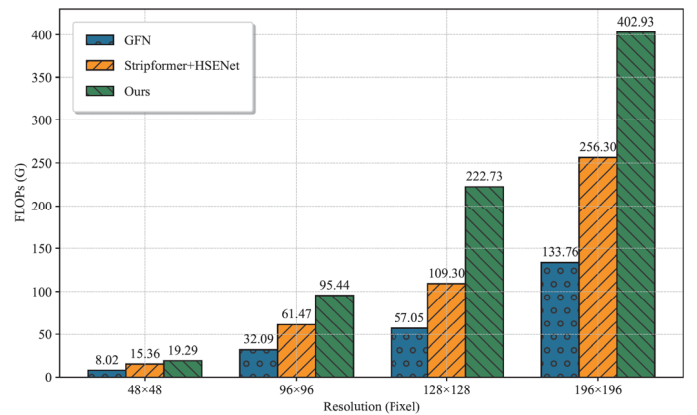

Fig. 12. FLOPs comparison of the GFN, Stripformer + HSENet, and AKMD-Net with diferent pixel resolutions.

*2) Effects on Remote Sensing Semantic Segmentation:* To validate the effectiveness of our proposed AKMD-Net method on high-level vision tasks, we further conduct experiments on the semantic segmentation task within the WHU Building Dataset. During the testing phase, we treat the original test data as high-quality images (HQ) and generate corresponding blurry LR images with a scale factor of 2. These blurry LR images are then reconstructed via multiple methods, including bicubic interpolation [49], RCAN [22], HAT [23], HSENet [24], Stripformer [21], GFN [12], Stripformer [21] + HSENet [24], HSENet [24] + Stripformer [21], and our proposed AKMD-Net. Subsequently, we evaluate the semantic segmentation results of the representative segmentation model segformer [76], which has been pretrained on the WHU building dataset. As shown in Table XII, images reconstructed by our AKMD-Net achieve the best semantic segmentation results based on commonly used metrics (mIoU, mAcc, and aAcc). It indicates that our method is capable of recovering more details than alternative approaches. Additionally, Fig. 13 in the response letter illustrates that the segformer [76] model achieves enhanced visual results when applied to images reconstructed by our AKMD-Net model, which highlights the importance of low-level algorithms in enhancing the performance of high-level tasks.

TABLE XI
ACCURACY RATES (%) ON THE IMAGES RECONSTRUCTED BY DIFFERENT LOW-LEVEL METHODS.

| Method | Top-1 Acc. | Top-5 Acc. |
|---|---|---|
| HQ | 96.51 | 99.78 |
| Bicubic [49] | 69.94 | 87.25 |
| RCAN [22] | 82.38 | 96.28 |
| HAT [23] | 81.21 | 95.42 |
| HSENet [24] | 81.51 | 95.78 |
| Stripformer [21] | 81.40 | 95.63 |
| GFN [12] | 81.37 | 95.56 |
| Stripformer [21] + HSENet [24] | 82.01 | 96.14 |
| HSENet [24] + Stripformer [21] | 81.95 | 96.09 |
| Ours | **82.73** | **96.54** |

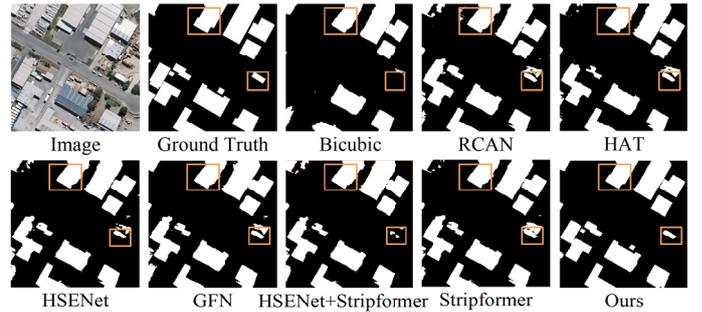

Fig. 13. Semantic segmentation results on the WHU Building dataset reconstructed by different low-level methods. Building: white and clutter: black.

## V. CONCLUSION

This paper presents a theoretical analysis of the spatial and blur image degeneration process and introduces a novel parallel dual-branch model named AKMD-Net to tackle the JRSIDSR task. By integrating the carefully designed PAKB and MDAB modules, the AKMD-Net is able to effectively extract high-quality features with various receptive fields. Besides, the AFF module is introduced to improve the cross-branch contextual relationships. At last, a novel AW Loss function is developed to depress the prior noise in the reconstructed images during training stage. Extensive experiments demonstrate the effectiveness of the proposed method. Although the proposed AKMD-Net achieves magnificent results on the JRSIDSR task, the way to fuse spatial and blur features can be further explored to adapt to more complex correlated degeneration and achieve better parameter efficiency.

TABLE XII
RESULTS OF THE DIFFERENT LOW-LEVEL METHODS ON THE HIGH-LEVEL TASK OF SEMANTIC SEGMENTATION. ALL SEMANTIC SEGMENTATION RESULTS ARE BASED ON THE SEGFORMER MODEL. THE BEST RESULTS ARE SHOWN IN BOLD.

| Method | mIoU↑ | mAcc↑ | aAcc↑ |
|---|---|---|---|
| HQ | 93.47 | 96.3 | 98.64 |
| Bicubic [49] | 79.31 | 83.77 | 95.59 |
| RCAN [22] | 90.13 | 94.57 | 97.88 |
| HAT [23] | 89.34 | 93.95 | 97.71 |
| HSENet [24] | 89.59 | 94.18 | 97.76 |
| Stripformer [21] | 89.42 | 94.17 | 97.72 |
| GFN [12] | 89.41 | 94.13 | 97.72 |
| Stripformer [21] + HSENet [24] | 89.70 | 94.20 | 97.79 |
| HSENet [24] + Stripformer [21] | 89.67 | 94.26 | 97.78 |
| Ours | **90.77** | **94.95** | **98.01** |

## REFERENCES

[1] Shijie Hao, Yuan Zhou, and Yanrong Guo. "A brief survey on semantic segmentation with deep learning". In: *Neurocomputing* 406 (2020), pp. 302–321.
[2] Zhengxia Zou et al. "Object detection in 20 years: A survey". In: *Proceedings of the IEEE* 111.3 (2023), pp. 257–276.
[3] Waseem Rawat and Zenghui Wang. "Deep convolutional neural networks for image classification: A comprehensive review". In: *Neural computation* 29.9 (2017), pp. 2352–2449.
[4] Ranganath R Navalgund, V Jayaraman, and PS Roy. "Remote sensing applications: An overview". In: *current science* (2007), pp. 1747–1766.
[5] Floyd F Sabins Jr and James M Ellis. *Remote sensing: Principles, interpretation, and applications*. Waveland Press, 2020.
[6] Shengyang Dai and Ying Wu. "Motion from blur". In: *2008 IEEE Conference on Computer Vision and Pattern Recognition*. IEEE. 2008, pp. 1–8.
[7] Kai Zhuang et al. "Multi-Domain Adaptation for Motion Deblurring". In: *IEEE Transactions on Multimedia* (2023).
[8] Alexander Shelekhov et al. "Low-altitude sensing of urban atmospheric turbulence with UAV". In: *Drones* 6.3 (2022), p. 61.
[9] Qiang Li et al. "Dual-stage approach toward hyperspectral image super-resolution". In: *IEEE Transactions on Image Processing* 31 (2022), pp. 7252–7263.
[10] Yi Xiao et al. "From degrade to upgrade: Learning a self-supervised degradation guided adaptive network for blind remote sensing image super-resolution". In: *Information Fusion* 96 (2023), pp. 297–311.
[11] Yujie Feng et al. "A Multiscale Generalized Shrinkage Threshold Network for Image Blind Deblurring in Remote Sensing". In: *IEEE Transactions on Geoscience and Remote Sensing* 62 (2024), pp. 1–16.
[12] Xinyi Zhang et al. "Gated fusion network for degraded image super resolution". In: *International Journal of Computer Vision* 128 (2020), pp. 1699–1721.
[13] Yong Li et al. "Pixel-level and perceptual-level regularized adversarial learning for joint motion deblurring and super-resolution". In: *Neural Processing Letters* 55.2 (2023), pp. 905–926.
[14] Xinyi Zhang et al. "A deep encoder-decoder networks for joint deblurring and super-resolution". In: *2018 IEEE international conference on acoustics, speech and signal processing (ICASSP)*. IEEE. 2018, pp. 1448–1452.
[15] Guanpeng Li et al. "Understanding error propagation in deep learning neural network (DNN) accelerators and applications". In: *Proceedings of the International Conference for High Performance Computing, Networking, Storage and Analysis*. 2017, pp. 1–12.
[16] Vijay Badrinarayanan, Alex Kendall, and Roberto Cipolla. "Segnet: A deep convolutional encoder-decoder architecture for image segmentation". In: *IEEE transactions on pattern analysis and machine intelligence* 39.12 (2017), pp. 2481–2495.

JOURNAL OF LATEX CLASS FILES, VOL. 18, NO. 9, SEPTEMBER 2020 15
[17] Estevão S Gedraite and Murielle Hadad. "Investigation on the effect of a Gaussian Blur in image filtering and segmentation". In: *Proceedings ELMAR-2011*. IEEE. 2011, pp. 393–396.
[18] Wenjie Luo et al. "Understanding the effective receptive field in deep convolutional neural networks". In: *Advances in neural information processing systems* 29 (2016).
[19] Ashish Vaswani et al. "Attention is all you need". In: *Advances in neural information processing systems* 30 (2017).
[20] Chao Dong et al. "Image super-resolution using deep convolutional networks". In: *IEEE transactions on pattern analysis and machine intelligence* 38.2 (2015), pp. 295–307.
[21] Fu-Jen Tsai et al. "Stripformer: Strip transformer for fast image deblurring". In: *European Conference on Computer Vision*. Springer. 2022, pp. 146–162.
[22] Yulun Zhang et al. "Image super-resolution using very deep residual channel attention networks". In: *Proceedings of the European conference on computer vision (ECCV)*. 2018, pp. 286–301.
[23] Xiangyu Chen et al. "Activating more pixels in image super-resolution transformer". In: *Proceedings of the IEEE/CVF conference on computer vision and pattern recognition*. 2023, pp. 22367–22377.
[24] Sen Lei and Zhenwei Shi. "Hybrid-scale self-similarity exploitation for remote sensing image super-resolution". In: *IEEE Transactions on Geoscience and Remote Sensing* 60 (2021), pp. 1–10.
[25] Sunghyun Cho and Seungyong Lee. "Fast motion deblurring". In: *ACM SIGGRAPH Asia 2009 papers*. 2009, pp. 1–8.
[26] Pawan Bharadwaj, Laurent Demanet, and Aimé Fournier. "Focused blind deconvolution". In: *IEEE transactions on signal processing* 67.12 (2019), pp. 3168–3180.
[27] Jian Sun et al. "Learning a convolutional neural network for non-uniform motion blur removal". In: *Proceedings of the IEEE conference on computer vision and pattern recognition*. 2015, pp. 769–777.
[28] Keiron O'shea and Ryan Nash. "An introduction to convolutional neural networks". In: *arXiv preprint arXiv:1511.08458* (2015).
[29] Ayan Chakrabarti. "A neural approach to blind motion deblurring". In: *Computer Vision–ECCV 2016: 14th European Conference, Amsterdam, The Netherlands, October 11-14, 2016, Proceedings, Part III 14*. Springer. 2016, pp. 221–235.
[30] Junyong Lee et al. "Iterative filter adaptive network for single image defocus deblurring". In: *Proceedings of the IEEE/CVF conference on computer vision and pattern recognition*. 2021, pp. 2034–2042.
[31] Syed Waqas Zamir et al. "Restormer: Efficient transformer for high-resolution image restoration". In: *Proceedings of the IEEE/CVF conference on computer vision and pattern recognition*. 2022, pp. 5728–5739.
[32] Zhendong Wang et al. "Uformer: A general u-shaped transformer for image restoration". In: *Proceedings of the IEEE/CVF conference on computer vision and pattern recognition*. 2022, pp. 17683–17693.
[33] James W Cooley, Peter AW Lewis, and Peter D Welch. "The fast Fourier transform and its applications". In: *IEEE Transactions on Education* 12.1 (1969), pp. 27–34.
[34] Jiwon Kim, Jung Kwon Lee, and Kyoung Mu Lee. "Accurate image super-resolution using very deep convolutional networks". In: *Proceedings of the IEEE conference on computer vision and pattern recognition*. 2016, pp. 1646–1654.
[35] Xin Li et al. "Learning omni-frequency region-adaptive representations for real image super-resolution". In: *Proceedings of the AAAI Conference on Artificial Intelligence*. Vol. 35. 3. 2021, pp. 1975–1983.
[36] Jie Hu, Li Shen, and Gang Sun. "Squeeze-and-excitation networks". In: *Proceedings of the IEEE conference on computer vision and pattern recognition*. 2018, pp. 7132–7141.
[37] Ze Liu et al. "Swin transformer: Hierarchical vision transformer using shifted windows". In: *Proceedings of the IEEE/CVF international conference on computer vision*. 2021, pp. 10012–10022.
[38] Pengcheng Zheng et al. "CGC-Net: A Context-Guided Constrained Network for Remote-Sensing Image Super Resolution". In: *Remote Sensing* 15.12 (2023), p. 3171.
[39] Yan Zhang et al. "FCIR: Rethink Aerial Image Super Resolution with Fourier Analysis". In: *ICASSP 2023-2023 IEEE International Conference on Acoustics, Speech and Signal Processing (ICASSP)*. IEEE. 2023, pp. 1–5.
[40] Xinyi Zhang et al. "A deep dual-branch networks for joint blind motion deblurring and super-resolution". In: *Proceedings of the 2nd International Conference on Vision, Image and Signal Processing*. 2018, pp. 1–6.
[41] Nour Aburaed et al. "A review of spatial enhancement of hyperspectral remote sensing imaging techniques". In: *IEEE Journal of Selected Topics in Applied Earth Observations and Remote Sensing* 16 (2023), pp. 2275–2300.
[42] Yi Qin et al. "Multi-Degradation Super-Resolution Reconstruction for Remote Sensing Images with Reconstruction Features-Guided Kernel Correction". In: *Remote Sensing* 16.16 (2024), p. 2915.
[43] Michael Elad and Arie Feuer. "Restoration of a single superresolution image from several blurred, noisy, and undersampled measured images". In: *IEEE transactions on image processing* 6.12 (1997), pp. 1646–1658.
[44] Junyi Li et al. "Spatially adaptive self-supervised learning for real-world image denoising". In: *Proceedings of the IEEE/CVF Conference on Computer Vision and Pattern Recognition*. 2023, pp. 9914–9924.
[45] Jérôme Lecoq et al. "Removing independent noise in systems neuroscience data using DeepInterpolation". In: *Nature methods* 18.11 (2021), pp. 1401–1408.
[46] Xingkai Yu and Jianxun Li. "Adaptive Kalman filtering for recursive both additive noise and multiplicative noise". In: *IEEE Transactions on Aerospace and Electronic Systems* 58.3 (2021), pp. 1634–1649.
[47] PC Hohenberg and JB Swift. "Effects of additive noise at the onset of Rayleigh-Bénard convection". In: *Physical Review A* 46.8 (1992), p. 4773.
[48] Sungchan Oh and Gyeonghwan Kim. "Robust estimation of motion blur kernel using a piecewise-linear model". In: *IEEE transactions on image processing* 23.3 (2014), pp. 1394–1407.
[49] Robert Keys. "Cubic convolution interpolation for digital image processing". In: *IEEE transactions on acoustics, speech, and signal processing* 29.6 (1981), pp. 1153–1160.
[50] Jingdong Chen et al. "New insights into the noise reduction Wiener filter". In: *IEEE Transactions on audio, speech, and language processing* 14.4 (2006), pp. 1218–1234.
[51] Orest Kupyn et al. "Deblurgan: Blind motion deblurring using conditional adversarial networks". In: *Proceedings of the IEEE conference on computer vision and pattern recognition*. 2018, pp. 8183–8192.
[52] Xin Tao et al. "Scale-recurrent network for deep image deblurring". In: *Proceedings of the IEEE conference on computer vision and pattern recognition*. 2018, pp. 8174–8182.
[53] Kaiming He et al. "Deep residual learning for image recognition". In: *Proceedings of the IEEE conference on computer vision and pattern recognition*. 2016, pp. 770–778.
[54] Shmuel Winograd. "On computing the discrete Fourier transform". In: *Mathematics of computation* 32.141 (1978), pp. 175–199.
[55] Chris Anderson and Marie Dillon Dahleh. "Rapid computation of the discrete Fourier transform". In: *SIAM Journal on Scientific Computing* 17.4 (1996), pp. 913–919.
[56] Wenzhe Shi et al. "Real-time single image and video super-resolution using an efficient sub-pixel convolutional neural network". In: *Proceedings of the IEEE conference on computer vision and pattern recognition*. 2016, pp. 1874–1883.
[57] Umme Sara, Morium Akter, and Mohammad Shorif Uddin. "Image quality assessment through FSIM, SSIM, MSE and PSNR—a comparative study". In: *Journal of Computer and Communications* 7.3 (2019), pp. 8–18.
[58] Justin Johnson, Alexandre Alahi, and Li Fei-Fei. "Perceptual losses for real-time style transfer and super-resolution". In: *Computer Vision–ECCV 2016: 14th European Conference, Amsterdam, The Netherlands, October 11-14, 2016, Proceedings, Part II 14*. Springer. 2016, pp. 694–711.
[59] Simonyan Karen. "Very deep convolutional networks for large-scale image recognition". In: *arXiv preprint arXiv: 1409.1556* (2014).
[60] Zhihao Wang, Jian Chen, and Steven CH Hoi. "Deep learning for image super-resolution: A survey". In: *IEEE transactions on pattern analysis and machine intelligence* 43.10 (2020), pp. 3365–3387.
[61] Qin Zou et al. "Deep learning based feature selection for remote sensing scene classification". In: *IEEE Geoscience and remote sensing letters* 12.11 (2015), pp. 2321–2325.
[62] Shunping Ji, Shiqing Wei, and Meng Lu. "Fully convolutional networks for multisource building extraction from an open aerial and satellite imagery data set". In: *IEEE Transactions on geoscience and remote sensing* 57.1 (2018), pp. 574–586.
[63] Zhiyuan Mao, Nicholas Chimitt, and Stanley H Chan. "Accelerating atmospheric turbulence simulation via learned phase-to-space transform". In: *Proceedings of the IEEE/CVF International Conference on Computer Vision*. 2021, pp. 14759–14768.
[64] Yujie Feng et al. "A Multiscale Generalized Shrinkage Threshold Network for Image Blind Deblurring in Remote Sensing". In: *IEEE Transactions on Geoscience and Remote Sensing* 62 (2024), pp. 1–16.
[65] Richard Zhang et al. "The unreasonable effectiveness of deep features as a perceptual metric". In: *Proceedings of the IEEE conference on computer vision and pattern recognition*. 2018, pp. 586–595.

[66] Hamid R Sheikh and Alan C Bovik. "A visual information fidelity approach to video quality assessment". In: *The first international workshop on video processing and quality metrics for consumer electronics*. Vol. 7. 2. sn. 2005, pp. 2117–2128.
[67] Xiaozhong Ji et al. "Real-world super-resolution via kernel estimation and noise injection". In: *proceedings of the IEEE/CVF conference on computer vision and pattern recognition workshops*. 2020, pp. 466–467.
[68] Rao Muhammad Umer, Gian Luca Foresti, and Christian Micheloni. *Deep Generative Adversarial Residual Convolutional Networks for Real-World Super-Resolution*. 2020. arXiv: 2005.00953 [eess.IV]. URL: https://arxiv.org/abs/2005.00953.
[69] Mekhak Shoyan et al. "Multi-stage progressive single image joint motion deblurring and super-resolution". In: *AIP Conference Proceedings*. Vol. 2757. 1. AIP Publishing. 2023.
[70] Wenqi Ren et al. "Gated fusion network for single image dehazing". In: *Proceedings of the IEEE conference on computer vision and pattern recognition*. 2018, pp. 3253–3261.
[71] Xingkai Yu and Jianxun Li. "Adaptive Kalman Filtering for Recursive Both Additive Noise and Multiplicative Noise". In: *IEEE Transactions on Aerospace and Electronic Systems* 58.3 (2022), pp. 1634–1649. DOI: 10.1109/TAES.2021.3117896.
[72] Reihaneh Malekian and Arian Maleki. "Is speckle noise more challenging to mitigate than additive noise?" In: *arXiv preprint arXiv:2409.16585* (2024).
[73] Satyakam Baraha and Ajit Kumar Sahoo. "Restoration of speckle noise corrupted SAR images using regularization by denoising". In: *Journal of Visual Communication and Image Representation* 86 (2022), p. 103546.
[74] Chongyi Li et al. "Low-Light Image and Video Enhancement Using Deep Learning: A Survey". In: *IEEE Transactions on Pattern Analysis and Machine Intelligence* 44.12 (2022), pp. 9396–9416. DOI: 10.1109/TPAMI.2021.3126387.
[75] Chengyang Li et al. "Detection-friendly dehazing: Object detection in real-world hazy scenes". In: *IEEE Transactions on Pattern Analysis and Machine Intelligence* 45.7 (2023), pp. 8284–8295.
[76] Enze Xie et al. "SegFormer: Simple and efficient design for semantic segmentation with transformers". In: *Advances in neural information processing systems* 34 (2021), pp. 12077–12090.

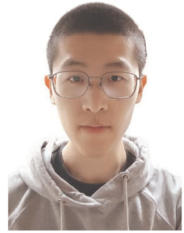

**Chengxiao Zeng** is an undergraduate student at Chongqing University of Posts and Telecommunications, Chongqing, China.

His research interests mainly include image reconstruction and deep learning.

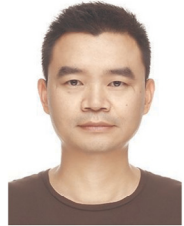

**Bin Xiao** received the B.S. and M.S. degrees in electrical engineering from Shanxi Normal University, Xi'an, China, in 2004 and 2007, respectively, and the Ph.D. degree in computer science from Xidian University, Xi'an, in 2012.

He is currently a Professor with the Chongqing University of Posts and Telecommunications, Chongqing, China. His research interests include image processing and pattern recognition.

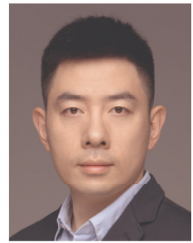

**Zhanghao Li** received the B.S. degree in telecommunications engineering and the Ph.D. degree in instrumentation science and technology from Chongqing University, China, in 2003 and 2009, respectively. In 2019, he has been with the Chongqing Institute of Green and Intelligent Technology, Chinese Academy of Sciences, where he is currently an Associate Professor.

His current research interests include computer vision and machine learning.

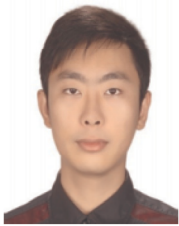

**Yan Zhang** received the B.S. degree from National University of Defense Technology, Changsha, China, in 2014, and the M.S. degree from the Univerity of Birmingham, United Kingdom, in 2017, and the Ph.D. degree from Chongqing University, Chongqing, China, in 2020.

He is currently a lecturer in Chongqing University of Posts and Telecommunications. His research area includes image processing, machine learning, and pattern recognition.

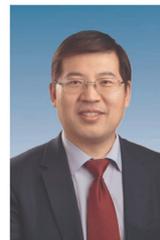

**Xinbo Gao** (M'02-SM'07-F'24) received the B.Eng., M.Sc. and Ph.D. degrees in electronic engineering, signal and information processing from Xidian University, Xi'an, China, in 1994, 1997, and 1999, respectively. From 1997 to 1998, he was a research fellow at the Department of Computer Science, Shizuoka University, Shizuoka, Japan. From 2000 to 2001, he was a post-doctoral research fellow at the Department of Information Engineering, the Chinese University of Hong Kong, Hong Kong. Since 1999, he has been at the School of Electronic Engineering, Xidian University and now he is a Professor of Pattern Recognition and Intelligent System of Xidian University. Since 2020, he has been also a Professor of Computer Science and Technology of Chongqing University of Posts and Telecommunications.

His current research interests include computer vision, machine learning and pattern recognition. He has published seven books and around 300 technical articles in refereed journals and proceedings. Prof. Gao is on the Editorial Boards of several journals, including Signal Processing (Elsevier) and Neurocomputing (Elsevier). He served as the General Chair/Co-Chair, Program Committee Chair/Co-Chair, or PC Member for around 30 major international conferences. He is Fellows of IEEE, IET, AAIA, CIE, CCF, and CAAI.

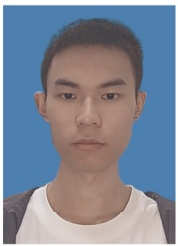

**Pengcheng Zheng** received the B.S. degree from Chongqing University of Posts and Telecommunications, ChongQing, China, in 2024. He is currently pursuing the M.S. degree with the School of Computer Science and Engineering, University of Electronic Science and Technology of China, Sichuan, China.

His research interests include image processing, large multimodal model, and transfer learning.